%% file: main.tex
\documentclass{article}

\input{math_commands.tex}

\usepackage{microtype}
\usepackage{graphicx}
\usepackage{subfigure}
\usepackage{makecell}
\usepackage{booktabs} %

\usepackage{enumitem}

\usepackage{hyperref}

\usepackage[accepted]{icml2025}

\usepackage{amsmath}
\usepackage{amssymb}
\usepackage{mathtools}
\usepackage{amsthm}

\usepackage{xcolor}
\usepackage{wrapfig}
\usepackage{multirow}
\usepackage{capt-of}

\usepackage[capitalize,noabbrev]{cleveref}

\usepackage{listings}
\definecolor{codegreen}{rgb}{0,0.6,0}
\definecolor{codegray}{rgb}{0.5,0.5,0.5}
\definecolor{codepurple}{rgb}{0.58,0,0.82}
\definecolor{backcolour}{rgb}{0.95,0.95,0.92}
\lstdefinestyle{mystyle}{
    backgroundcolor=\color{backcolour},   
    commentstyle=\color{codegreen},
    keywordstyle=\color{magenta},
    numberstyle=\tiny\color{codegray},
    stringstyle=\color{codepurple},
    basicstyle=\ttfamily\footnotesize,
    breakatwhitespace=false,         
    breaklines=true,                 
    captionpos=b,                    
    keepspaces=true,                 
    numbers=left,                    
    numbersep=5pt,                  
    showspaces=false,                
    showstringspaces=false,
    showtabs=false,                  
    tabsize=2
}
\theoremstyle{plain}

\theoremstyle{definition}

\theoremstyle{remark}

\usepackage[textsize=tiny]{todonotes}

\newcommand*{\Perm}[2]{{}^{#1}\!P_{#2}}%
\newcommand{\X}{\mathbf{X}}
\newcommand{\until}{\mathbf{U}}

\icmltitlerunning{Interchangeable Token Embeddings for Extendable Vocabulary and Alpha-Equivalence}

\begin{document}

\twocolumn[
\icmltitle{Interchangeable Token Embeddings for Extendable Vocabulary and Alpha-Equivalence}

\begin{icmlauthorlist}
\icmlauthor{İlker Işık}{yyy}
\icmlauthor{Ramazan Gokberk Cinbis}{yyy}
\icmlauthor{Ebru Aydin Gol}{comp}
\end{icmlauthorlist}

\icmlaffiliation{yyy}{Department of Computer Engineering, Middle East Technical University, Ankara, Turkey}
\icmlaffiliation{comp}{Microsoft, İstanbul, Turkey}

\icmlcorrespondingauthor{İlker Işık}{ilker@ceng.metu.edu.tr}
\icmlcorrespondingauthor{Ramazan Gokberk Cinbis}{gcinbis@ceng.metu.edu.tr}
\icmlcorrespondingauthor{Ebru Aydin Gol}{ebruaydingol@microsoft.com}

\icmlkeywords{Machine Learning, ICML}

\vskip 0.3in
]

\printAffiliationsAndNotice{}  %

\begin{abstract}
\input{sections/abstract}
\end{abstract}

\input{sections/intro}
\input{sections/related}  %
\input{sections/problem}  %
\input{sections/method}

\input{sections/experiments}
\input{sections/limitations}

\input{sections/conclusion}

\section*{Acknowledgments}
The numerical calculations were partially performed at TÜBİTAK TRUBA, MareNostrum5, METU ImageLab, and METU ROMER resources. This project was supported in part by the project METU ADEP-312-2024-11525. Dr. Cinbis is supported by the ``Young Scientist Awards Program (BAGEP)‘’ of Science Academy, Türkiye.

\input{sections/impact}

\bibliography{main}
\bibliographystyle{icml2025}

\newpage

\appendix

\onecolumn

\input{sections/prelim}
\input{sections/ltl}
\input{sections/prop}
\input{sections/hyperparam}

\input{sections/copying}
\input{sections/limited}

\input{sections/propexp}
\input{sections/efficiency}
\input{sections/llama}

\end{document}

%% file: math_commands.tex
\usepackage{amsmath,amsfonts,bm}

\def\eqref#1{equation~\ref{#1}}
\def\Eqref#1{Equation~\ref{#1}}

\def\1{\bm{1}}

\def\rva{{\mathbf{a}}}

\def\erva{{\textnormal{a}}}

\def\va{{\bm{a}}}
\def\vb{{\bm{b}}}

\def\vv{{\bm{v}}}

\def\vx{{\bm{x}}}
\def\vy{{\bm{y}}}

\def\valpha{{\bm{\alpha}}}
\def\vbeta{{\bm{\beta}}}

\def\eva{{a}}
\def\evb{{b}}

\def\mL{{\bm{L}}}

\def\mU{{\bm{U}}}

\def\mX{{\bm{X}}}

\DeclareMathAlphabet{\mathsfit}{\encodingdefault}{\sfdefault}{m}{sl}
\SetMathAlphabet{\mathsfit}{bold}{\encodingdefault}{\sfdefault}{bx}{n}

\def\sP{{\mathbb{P}}}

\def\sR{{\mathbb{R}}}

\def\sU{{\mathbb{U}}}
\def\sV{{\mathbb{V}}}

%% file: sections/abstract.tex
Language models lack the notion of interchangeable tokens: symbols that are semantically equivalent yet distinct, such as bound variables in formal logic.
This limitation prevents generalization to larger vocabularies and hinders the model's ability to recognize alpha-equivalence, where renaming bound variables preserves meaning.
We formalize this machine learning problem and introduce alpha-covariance, a metric for evaluating robustness to such transformations.
To tackle this task, we propose a dual-part token embedding strategy: a shared component ensures semantic consistency, while a randomized component maintains token distinguishability.
Compared to a baseline that relies on alpha-renaming for data augmentation, our approach demonstrates improved generalization to unseen tokens in linear temporal logic solving, propositional logic assignment prediction, and copying with an extendable vocabulary, while introducing a favorable inductive bias for alpha-equivalence.
Our findings establish a foundation for designing language models that can learn interchangeable token representations, a crucial step toward more flexible and systematic reasoning in formal domains.
Our code and project page are available at \href{https://necrashter.github.io/interchangeable-token-embeddings}{necrashter.github.io/interchangeable-token-embeddings}

%% file: sections/intro.tex
\section{Introduction}
\label{sec:intro}

Following the deep learning revolution that affected numerous application areas \citep{dlSurvey2020},
recent literature shows that deep learning based approaches also perform well in neurosymbolic reasoning tasks, such as theorem proving \citep{Han2021ProofAC} and mathematical reasoning \citep{Rabe2020MathematicalRV}.
The formal reasoning capabilities of these models were once doubted, but \citet{liu2023transformers} demonstrated the ability of Transformer models \citep{vaswani} to learn shortcuts to automata.
Of particular interest is the generalization ability of such models to unseen, out-of-distribution data \citep{Sanh2021MultitaskPT}, enhancing their appeal for logical reasoning \citep{pmlr-v202-abbe23a}.

Another application area is linear-time temporal logic (LTL), which is heavily utilized by the formal verification community \citep{Clarke2018HandbookOM,Baier2008PrinciplesOM} for reasoning about how logical propositions change over time~\citep{Pnueli77}.
Through the use of temporal operators, LTL formulae can specify, for example, that a proposition $p$ must hold at all time steps ($\mathbf{G} p$), or at least one time step ($\mathbf{F} p$).
LTL formulae operate on traces, which describe how the propositions change over time.

Solving a given LTL formula involves finding a satisfying trace, and it proved essential for generating examples for system specifications in the literature.
This field was dominated by the methods that use classical algorithms, such as \verb|spot| \citep{spot} and \verb|aalta| \citep{aalta}.
However, following the success of Transformer models on end-to-end symbolic integration \citep{Lample2019DeepLF}, \citet{deepltl} attacked the LTL solving problem using the same approach. %
Their capability to generalize to longer formulae is especially noteworthy, and it was made possible thanks to tree-positional encoding \citep{treepos}.

However, generalization to longer formula lengths is not the only concern.
In particular, each LTL formula features a set of atomic propositions (henceforth APs), and it's desirable for the model to generalize to more APs.
But the architecture of the model does not even accept new APs that are not seen during training, despite the fact that all APs represent \textit{semantically equivalent} concepts while being \textit{distinguishable} from each other.
This situation arises in many other application areas, such as mathematical expressions and lambda calculus \citep{alpha-conversion}, where renaming the bound variables does not change the meaning.
This phenomenon is described as \textit{alpha-equivalence}. %
\textit{Alpha-conversion} (or \textit{alpha-renaming}) refers to the process of creating alpha-equivalent input-output pairs.

In this paper, we propose a novel approach for representing interchangeable tokens in neural network models.
To summarize, our method constructs some part of the token embeddings on-the-fly instead of learning all of them during training.
The token embeddings for interchangeable tokens consist of two parts: a learnable part and a randomized part.
The learnable part is shared across all interchangeable tokens, and the model must depend on the randomized part to differentiate these tokens.
Thanks to the randomized component, our method can generate embeddings for arbitrarily many interchangeable tokens as needed during both training and inference, with the only practical limitation being the exponentially growing sampling set size for discrete random generation methods.
We use the weight tying technique \citep{weight-tying} to share the same token embeddings with the final projection matrix, which calculates the logits (i.e., next-token probabilities before softmax).

We use our embedding method in a Transformer encoder-decoder model and evaluate it on three tasks: copying with an extendable vocabulary, solving LTL formulae, and predicting assignments for propositional logic.
As a baseline, we consider a simpler approach that uses alpha-renaming for data augmentation during training to expose the model to a larger vocabulary, which is also new in the literature to the best of our knowledge.
Overall, our method demonstrates generalization capabilities to larger vocabulary sizes, and also combines well with positional encodings that exhibit length generalization.
We also experiment with dataset perturbation to show that our method introduces a helpful inductive bias for alpha-equivalence.
Finally, we present \textit{alpha-covariance}, a metric for measuring robustness against alpha-conversions that is applicable to any domain where alpha-equivalence is relevant. %

Overall, our contributions can be summarized as follows.
\begin{enumerate}[itemsep=-3pt,topsep=-1pt,leftmargin=12pt]
    \item Identify the problem of generalizing to larger vocabularies in (formal) language modeling tasks, and define an experimental protocol to study this problem.
    \item Propose alpha-covariance, a novel metric for measuring robustness against alpha-conversions, applicable to any domain with interchangeable tokens.
    \item Introduce a dual-part embedding method for vocabulary generalization and improved alpha-covariance, with negligible computational overhead.
    \item Verify the proposed method thoroughly on three tasks: copying with extendable vocabulary, solving LTL formulae, and predicting assignments for propositional logic.
\end{enumerate}

%% file: sections/related.tex
\section{Related Work}
\label{sec:related}

\textbf{Language modeling and formal reasoning.}
The transformer architecture \citep{vaswani}, now ubiquitous in modern deep learning, was initially proposed as a generative model to translate between natural languages autoregressively.
This led to many successful attempts to frame formal reasoning tasks as language modeling problems, such as symbolic integration \citep{Lample2019DeepLF}, symbolic regression \citep{EndtoendSR,symformer}, LTL solving \citep{deepltl}, and many more.
Further developments shifted the field towards large language models (LLMs), e.g.,
by prompting a model pre-trained on a gigantic scale \citep{chatgpt-math},
by enhancing the prompt with retrieved references for proof generation \citep{NaturalProver,LeanDojo},
by training an LLM on a specialized dataset for mathematics \citep{LlemmaModel}.
However, the reasoning abilities of LLMs were questioned by \citep{Tang2023LargeLM}, who showed LLMs struggle with symbolic reasoning when semantics are decoupled, and by others \citep{reasoning-or-reciting}.

\textbf{Extensible vocabulary.}
Efforts to create an extensible vocabulary for neural networks are scarce in the broader machine learning community, let alone the formal reasoning literature.
\citet{def2vec} exploited dictionary definitions to create extensible word embeddings.
\citet{sign-lang} proposed a vocabulary-extensible sign language recognition framework by using a component based approach, where each sign gesture is recognized based on common components such as hand shape, orientation, axis, rotation, and trajectory.
These studies depend on either external information (dictionary definitions) or properties specific to an application area (components of hand gesture); they do not attempt to design an extensible vocabulary for interchangeable tokens, which has been neglected by the literature alongside the concept of alpha-equivalence.

%% file: sections/problem.tex
\section{Problem Definition}

In language modeling, the goal is to predict the next token in the output sequence given the input and the past output.
(See Appendix \ref{app:llm-prelim} for more background.)
Let $\sV$ denote the set of all unique tokens, i.e., the vocabulary of a language modeling problem.
We use $\sV^*$ to denote the set of all finite sequences of tokens (strings) from $\sV$.
We assume that $\sV_i$ is the set of interchangeable tokens and $\sV_n = \sV \backslash \sV_i$ is the set of non-interchangeable tokens.
The core idea behind \textit{alpha-equivalence} is that renaming interchangeable tokens between each other in both input and output preserves meaning.
Let $f \colon \sV \to \sV$ be a bijection such that $f(x) = x$ for all $x \in \sV_n$, i.e., $f$ arbitrarily renames the interchangeable tokens between each other in one-to-one correspondence and preserves the rest of the tokens.
We apply $f$ to each token in a given pair of input $\va \in \sV^*$ and output $\vb \in \sV^*$ strings, obtaining $\va' = (f(\eva_1), f(\eva_2), \ldots)$ and $\vb' = (f(\evb_1), f(\evb_2), \ldots)$.
We call this operation \textit{alpha-conversion} or \textit{alpha-renaming}.
The set of interchangeable tokens $\sV_i$ must be defined such that $\va'$ and $\vb'$ form a valid input-output pair semantically equivalent to $(\va, \vb)$ for all possible $f$.

Our task is to design an embedding method that---alongside being resilient to alpha-renaming by construction---can support a new vocabulary $\sV' = \sV_i' \cup \sV_n$ where $\sV_i \subset \sV_i'$ after training on $\sV$.
In other words, the model should be able to operate on a larger vocabulary than the one seen during training, as long as the newly introduced tokens belong to the class of interchangeable tokens.
Although we don't impose any restrictions about the size of $\sV'$ in this problem definition,
the maximum size of $\sV'$ in practice may change as a function of the number of embedding dimensions.
Thus, while setting the hyperparameters, the expected size of $\sV'$ must be considered.

\textbf{Example.}
In the LTL solving problem (Appendix \ref{app:ltl}),
the set of non-interchangeable tokens $\sV_n$ includes the operators, constants, delimiter tokens (``\verb|;|'', ``\verb|{|'', ``\verb|}|''), and any special tokens such as the end token.
The set of interchangeable tokens equals to the set of atomic propositions (APs): $\sV_i = P$.
Assuming $P = \{\texttt{a}, \texttt{b}\}$, the formula-trace pair (``\verb|&aXb|'', ``\verb|a;b;{1}|'') is alpha-equivalent to (``\verb|&bXa|'', ``\verb|b;a;{1}|'').
Further, assume that the augmented set of interchangeable tokens is $\sV_i' = P' = \{\texttt{a}, \texttt{b}, \texttt{c}, \texttt{d}\}$.
Now, the aforementioned pair can also be equivalently represented as (``\verb|&cXd|'', ``\verb|c;d;{1}|'').
The augmented vocabulary allows the expression of formula-trace pairs that feature up to 4 APs instead of 2.
For example, (``\verb|&&abX&cd|'', ``\verb|&ab;&cd;{1}|'') cannot be expressed using $P = \{\texttt{a}, \texttt{b}\}$.
Our goal is to create a model that can handle such inputs despite being trained on the limited vocabulary $\sV = \sV_n \cup P$.

%% file: sections/method.tex
\section{Proposed Method}
\label{sec:method}

\begin{figure*}
  \begin{minipage}{.3\textwidth}
    \centering
    \includegraphics[width=.9\linewidth]{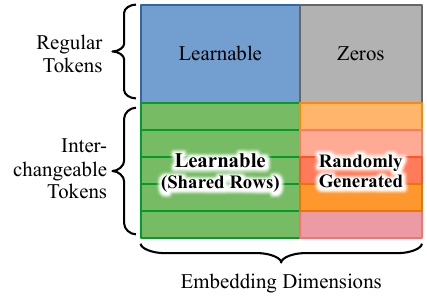}
    \caption{
      Visual structure of the embedding matrix in the proposed method.
    }
    \label{fig:embed}
  \end{minipage} \quad
  \begin{minipage}{.69\textwidth}
      \centering
      \vspace{-0.2in}
      \captionof{table}{Comparison of random vector generation methods.}
      \label{random-methods-table}
      \vspace{0.1in}
      \scalebox{0.85}{
      \begin{tabular}{l|ccc}
          {\bf Method}  & {\bf Normal Distribution}  & {\bf Neighboring Points}  & {\bf Hypercube Vertices} \\
          \hline
          \multirow{2}{*}{\textbf{Formula}}
          & $\erva_i \sim \mathcal{N}(0, 1)$
          & $\erva_i \in \{ -1, 0, 1\}$
          & $\erva_i \in \{ -1, 1\}$
          \\
          & 
          & $\|\rva\| \neq 0$
          & 
          \\
          \hline
          \textbf{Size} for $n$-dims
          & Continuous
          & $3^n - 1$
          & $2^n$
          \\
          \hline
          \textbf{Sample Visualization}
          & \raisebox{-0.5\totalheight}{\includegraphics[width=2cm]{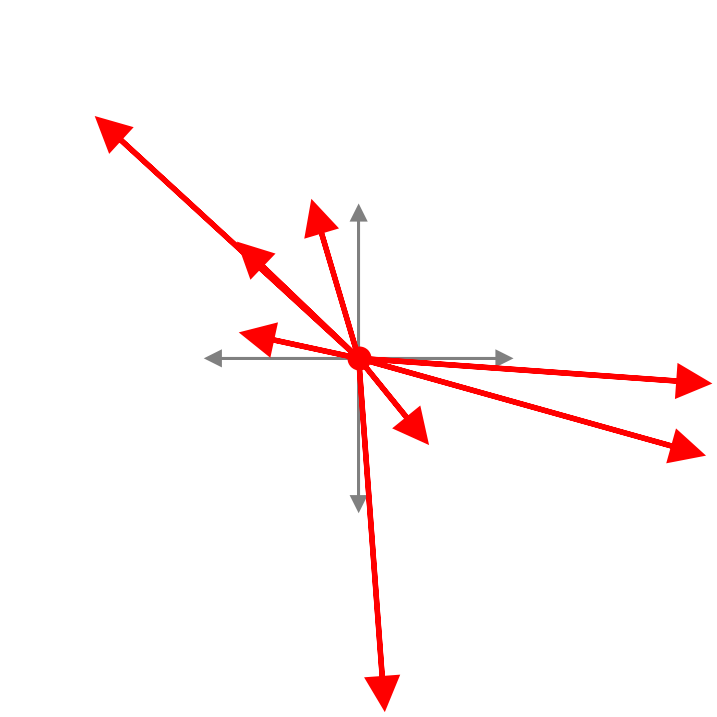}}
          & \raisebox{-0.5\totalheight}{\includegraphics[width=2cm]{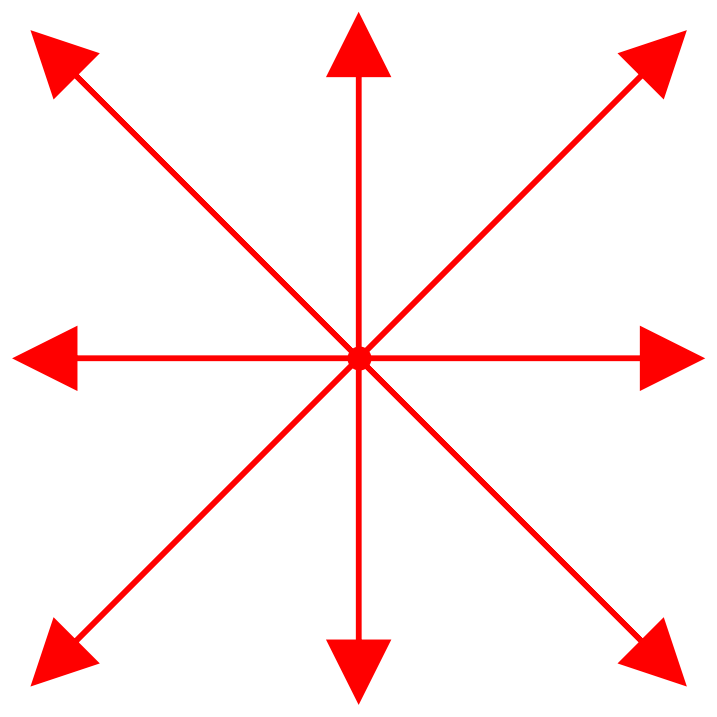}}
          & \raisebox{-0.5\totalheight}{\includegraphics[width=2cm]{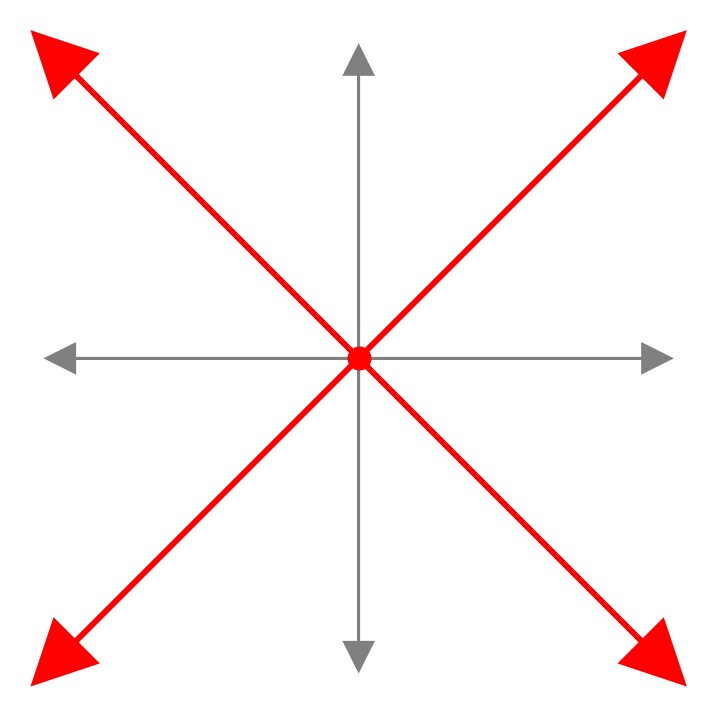}}
          \\
      \end{tabular}
      }
  \end{minipage}
\end{figure*}

To address the problem of learning semantically equivalent but distinguishable (alpha-equivalent) tokens, our method employs two ideas:
sharing some part of the embeddings between such tokens to convey their semantic equivalence;
and  assigning a unique randomly-generated vector to the rest of the embedding for each interchangeable token, allowing the model to distinguish between them.
The number of shared and randomly-generated dimensions are denoted by $d_\alpha$ and $d_\beta$ respectively.
The sum of these two yields the total number of embedding dimensions in the model, denoted by $d_\text{model} = d_\alpha + d_\beta$.
For non-interchangeable tokens, $d_\alpha$ dimensions contain separate learnable parameters and $d_\beta$ dimensions are set to $0$.
The structure of the embedding matrix is visualized in Figure \ref{fig:embed}.

\subsection{Embedding matrix}
\label{sec:method:embed}

\textbf{Construction of the embedding matrix.}
For a vocabulary with $n$ non-interchangeable tokens and $m$ interchangeable tokens,
$\mL \in \sR^{n \times d_\alpha}$ represents the matrix of learnable embeddings for non-interchangeable tokens,
$\valpha \in \sR^{1 \times d_\alpha}$ the shared learnable embedding for interchangeable tokens,
and $\vbeta_i \in \sR^{1 \times d_\beta}$ the randomly-generated embedding for the $i$th interchangeable token where $1 \leq i \leq m$.
Note that $\valpha$ and $\vbeta_i$ are row vectors.
A zero matrix of size $i \times j$ is represented by $\bm{0}^{i,j}$.
In addition, we define two row-based L2 normalization functions $f_{bn}(\mX)$ and $f_{fn}(\mX)$ that divide each row $\mX_{i,:}$ by its L2 norm $\| \mX_{i,:} \|$.
These two functions are identical but can be disabled independently from each other, hence the separation.
Finally, the overall structure of the embedding matrix $\mU$ is shown in \Eqref{eq:embed}.
In this construction, the interchangeable tokens are assumed to come after the non-interchangeable tokens.
Note that it's also possible to implement multiple sets of different interchangeable tokens via a trivial extension.

\begin{equation}
  \label{eq:embed}
  \mU = f_{fn} (\begin{bmatrix}
    f_{bn}(\mL) && \bm{0}^{n, d_\beta} \\
    f_{bn}(\valpha) && f_{bn}(\vbeta_1) \\
    f_{bn}(\valpha) && f_{bn}(\vbeta_2) \\
    & \vdots & \\
    f_{bn}(\valpha) && f_{bn}(\vbeta_m) \\
  \end{bmatrix})
\end{equation}

During training, the embedding matrix must be reconstructed in each forward pass with resampled random vectors $\vbeta_1$ to $\vbeta_m$.
Resampling $\vbeta_i$ for $1 \leq i \leq m$ during training prevents the model from adapting to the idiosyncracies of a particular random generation and forces it to distinguish between interchangeable tokens regardless of the contents of $\vbeta_i$.
During inference, it's created once at the start and remains the same since the autoregressive generation involves multiple forward passes on the same input.

\textbf{Normalization.}
There are several concerns that warrant the heavy use of normalization while constructing $\mU$, as seen in \Eqref{eq:embed}.
Firstly, $d_\alpha$ dimensions and $d_\beta$ dimensions should not overwhelm each other in terms of magnitude.
Normalizing $\valpha$ and $\vbeta_i$ separately addresses this issue.
The magnitude of the concatenated embedding is another concern, which is handled by the final normalization.
The normalization of $\mL$ is redundant (since the final normalization does the same operation after the concatenation with zeros) but kept in \Eqref{eq:embed} for readability.

\subsection{Random embedding generation}
\label{sec:method:random}

This section will explain how the distinguishing part of the interchangeable token embeddings, $\vbeta_i, 1 \leq i \leq m$, are created.
To this end, we developed 3 methods to generate random vectors.
Table \ref{random-methods-table} provides a summary at a glance.
The first method simply samples the standard normal distribution for each dimension.
The second one uses the neighboring grid points around the origin, which correspond to the 8 directions in 2D.
For each interchangeable token, a unique vector in this set is sampled.
The last method is similar, but its set consists of the vertices of a hypercube centered around the origin, i.e., diagonal direction vectors.

\textbf{Uniqueness constraint.}
In the normal distribution method, we don't have any additional constraints to ensure distinguishability between vectors.
However, in other two methods, we need to make sure that each interchangeable token gets assigned to a unique vector since the sampling set is finite.
To achieve this quickly and space-efficiently, we define a mapping from integers to possible vectors.
The unique vectors are generated by sampling $m$ unique random integers (which can be calculated efficiently using the reservoir sampling technique),
and then using the defined mapping to convert these integers to the vectors.
This strategy avoids materializing the whole set of possible vectors.
In the hypercube vertices method, we map the binary digits of an integer in $[0, 2^{d_\beta})$ to $\{-1, 1\}$.
Although ``Neighboring Points" is simply the ternary version of the same idea, avoiding the zero vector requires special care.
The zero vector maps to the integer $i_z = (3^{d_\beta}-1)/2$.
Therefore, we define our domain as the integers in $[0, 3^{d_\beta} - 1)$ and add $1$ to the integer $i$ before converting it if $i \geq i_z$.
Integer mapping approach for generating unique vectors works well for up to 32 dimensions, after which the limits of integer representation become an issue for reservoir sampling.
Therefore, in such cases, we simply disable the uniqueness check because the exponentially growing size of the sampling set renders the probability of drawing the same sample negligible.

\subsection{Projection}
\label{sec:method:project}

\textbf{Weight tying.}
In a traditional language modeling setting, since both the embedding and projection matrices are entirely composed of learnable parameters, it's not necessary to share them, even though there are many advantages of weight tying \citep{weight-tying}.
However, we construct the embedding matrix manually in our method, which makes weight tying a requirement.
Furthermore, since we perform our experiments on an encoder-decoder architecture in this paper, we utilize a three-way weight tying approach, whereby the embedding matrices of encoder and decoder are tied in addition to the final projection matrix.
Three-way weight tying is particularly appropriate for the LTL solving task since many tokens are shared between the LTL formulae and traces.

\textbf{Feature normalization.}
Given the output of the last layer before the final projection $\vv$ (henceforth called feature vector), instead of directly applying the final projection as in $\mU \vv$, we apply L2 normalization to the feature vector $\vv$ before passing it through the final projection: $\mU f_{fn}(\vv)$.
This matrix multiplication constitutes taking a dot product with each row.
Since $\va \cdot \vb = \|\va\| \|\vb\| cos(\theta)$ where $\theta$ is the angle between $\va$ and $\vb$,
normalizing both the embeddings and the feature vector %
leaves only the cosine term to determine the logits.
This forces the model to distinguish between tokens based solely on the directions, which may improve the gradient flow.

\textbf{Cosine loss.}
If we normalize both the embeddings and the feature vector, the only thing that determines each logit is the cosine of the angle between the feature vector and the embedding.
Applying the softmax loss to such logits is known as cosine loss in the literature.
Although cosine-based loss functions were successful in face recognition \citep{l2face, normface}, it proved sensitive to hyperparameter settings in these losses.
To avoid this problem, we use AdaCos loss function \citep{adacos} that scales the logits adaptively throughout training.

Despite the attractiveness of AdaCos in this context, it is not directly applicable in a language modeling setting due to the additional sequence length dimension,
and no prior work explored this application to the best of our knowledge.
To overcome this, we modify the AdaCos loss function as follows:
First, we combine the batch and length dimensions while ignoring the padding tokens, effectively treating both dimensions as batch dimensions.
However, since this change greatly increases the number of batch dimensions, it can lead to numerical issues, even with the log-sum-exp trick.
Therefore, we clip the scale value calculated by AdaCos to a maximum of 100 to avoid numerical issues.
This loss formulation can also be used with conventional embeddings, as we do in our experiments.

%% file: sections/experiments.tex
\begin{figure*}
  \centering
  \includegraphics[width=\textwidth]{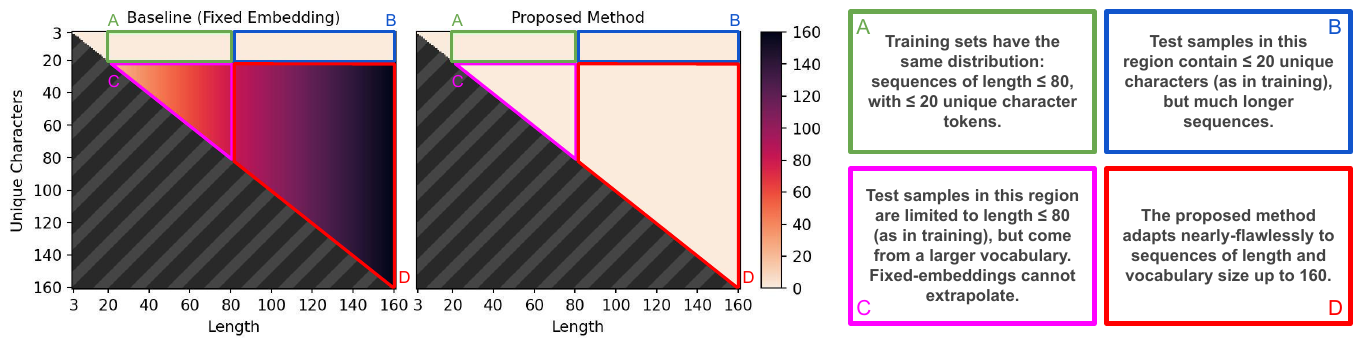}
  \caption{ 
    Two annotated heatmaps visualizing the test-set edit distance between prediction and ground truth in copying task with extendable vocabulary.
    Both heatmaps share the same y-axis.
    The green box represents the number of unique characters (y-axis) and the maximum length (x-axis) in the training dataset.
    Each point shows the average test error, except the lower triangular part of each heatmap (gray hatch pattern) corresponding to the impossible combinations of length and unique character counts.
    The traditional approach (left), using ubiquitously utilized fixed (learned) token embeddings, cannot extrapolate to vocabulary expansions.
    The proposed method (right) enables generalization to larger vocabulary sizes at longer sequence lengths, compared to what is observed during training.
    }
  \label{fig:intro-result}
\end{figure*}

\section{Experiments}
\label{sec:experiments}

\textbf{Experimental setup.}
We use a transformer encoder-decoder architecture in all experiments.
We always use the same embedding size in both encoder and decoder due to weight tying.
We use the RoPE~\citep{rope} as the positional encoding method in the decoder.
In the encoder, we use tree-positional encoding if applicable (logic tasks), RoPE otherwise (copying task).
The hyperparameter settings are given in Table~\ref{hyperparam-table} in Appendix~\ref{app:hyperparam}.

\textbf{Baselines.}
We train three types of baseline models with traditional embeddings: the first one on the original dataset, the second one on a dataset with the same parameters but using a larger vocabulary size, and the third one on the original dataset but using a data augmentation strategy.
Specifically, for the third baseline, the number of interchangeable token embeddings matches that of the test set, and we apply random alpha-renaming at each forward pass during training.
This ensures that the model is exposed to all tokens in the test set, but the number of unique interchangeable tokens the model sees in each sample remains limited as in the training set.
Note that this is an internal baseline that doesn't exist in the literature to the best of our knowledge.

\subsection{Copying with Extendable Vocabulary}
\label{sec:experiments:copying}

We introduce a new toy problem designed to evaluate the vocabulary generalization capabilities of our embedding method.
We create various training datasets that contain 10 million random strings with a limited vocabulary size.
A string is given as input, and the model is expected to produce the input string exactly via autoregressive generation.
This embodies a helpful toy problem for our method because all tokens are interchangeable, barring the special tokens (start/end).
In these experiments, we expect the model to generalize to larger vocabulary sizes unseen during training.

Using edit distance as our evaluation metric, we first assess the vocabulary generalization capabilities (Appendix \ref{sec:copy-1}).
Since our method excels in this task, we then explore generalization in both vocabulary size and string length (Appendix \ref{sec:copy-2}), performing a hyperparameter search over the settings of our embedding method (Appendix \ref{app:copying:hyperparam-search}).
Finally, we scale up the vocabulary size and the string lengths to evaluate our method (Appendix \ref{sec:copy-big}).
Our method exhibits perfect performance in the out-of-distribution domain as shown in Figure \ref{fig:intro-result}.
We also examine our method's sensitivity to randomness in embeddings (Appendix \ref{sec:copy-randomness-sensitivity}), and propose using the random embedding with median cross entropy loss as a proxy for average performance.

\subsection{LTL Solving}
\label{sec:experiments:ltl}

In this section, we train models on the LTLRandom35 dataset from DeepLTL \citep{deepltl} and other synthetic datasets created with the same method.
To evaluate the correctness of the generated formulae, we utilize \verb|spot| framework version \verb|2.11.6| \citep{spot}.
We use tree-positional encoding \citep{treepos} in the encoder and RoPE \citep{rope} in the decoder.
We generate predictions using beam search with beam size = 3.

\input{sections/table_alpha_cov.tex}

\textbf{Baselines.}
We trained all of the baseline models from scratch.
For the first type of baseline, we aimed to reproduce the results from \citet{deepltl}.
Hence, we used the best hyperparameters they reported (Appendix \ref{app:hyperparam}).
Unlike \citet{deepltl}, we experimented with RoPE (in the decoder) and AdaCos, but did not observe a noteworthy improvement on the validation set.\footnotemark
After determining the best baseline model on the validation set, %
we evaluated it on the test split of LTLRandom35 and obtained a correct rate of $98.2\%$ against the $98.5\%$ reported by \citet{deepltl}.

\footnotetext{Using RoPE in the decoder increased the ratio of correct predictions from 97.8\% to 98.0\% on the validation set. Introducing AdaCos in addition to RoPE increased this value to 98.2\%.}

\subsubsection{Dataset Perturbations}
\label{sec:perturbation}

To demonstrate that our method creates a helpful inductive bias, we created a perturbed version of the LTLRandom35 dataset by renaming the APs such that the order of the first AP appearances in the trace is always the same.
As the empirical evidence in Table \ref{perturbation-table} confirms, both our method and the alpha-renaming baseline are naturally immune to these alterations.
We train these methods only on the perturbed dataset since training them again on the normal dataset amounts to training with different random samples.

While the original model performs significantly worse under perturbation,
both alpha-renaming and proposed models match the baseline performance in correctness ratio despite perturbation.
This observation suggests that these modifications introduce a robust inductive bias that makes the models resistant to perturbations in the data.
A minor decrease in the ratio of exact matches is noted, but this may signify less overfitting and a better bias-variance tradeoff in the larger context.
Appendix \ref{app:ltl-limited} continues this experiment with limited amount of training samples instead of perturbations.

\begin{figure*}[p]
  \centering
    \subfigure[LTL Solving]{\label{fig:ltl-heatmap}\includegraphics[width=\textwidth]{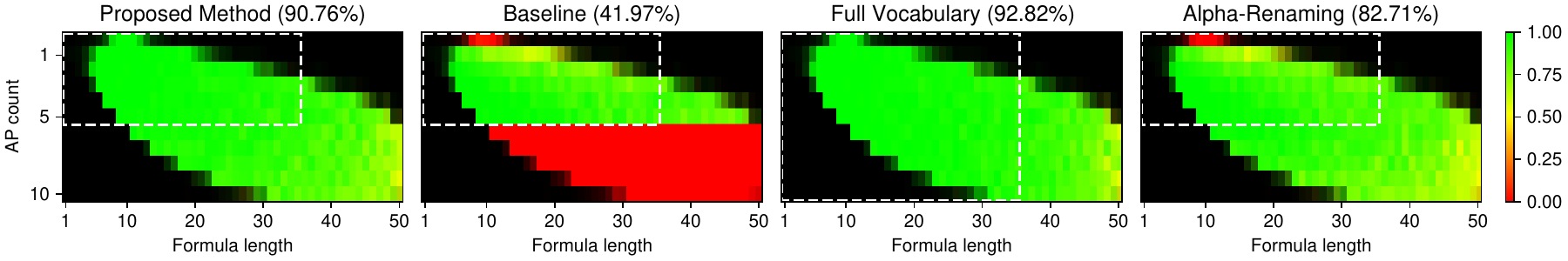}}
    \subfigure[Propositional Logic]{\label{fig:prop-heatmap}\includegraphics[width=\textwidth]{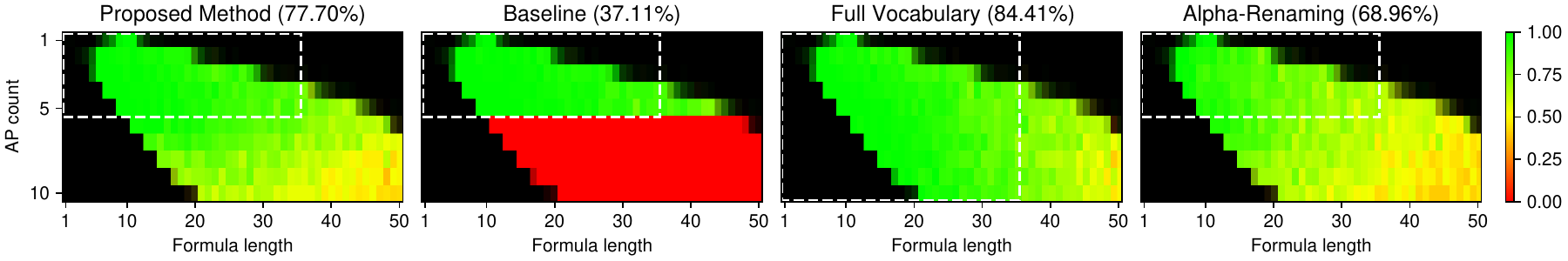}}
  \caption{
    Heatmaps visualizing the ratio of correct predictions on a special test set, for LTL solving (top) and propositional logic (bottom) tasks.
    The brightness of the color depends on the sample size, with full brightness representing 100 samples.
    The dashed white box represents the boundaries of the training dataset.
    Our model is competitive with the full vocabulary baseline despite being only trained on formulae with at most 5 APs, and outperforms other baselines.
  }
  \label{fig:heatmaps}
\end{figure*}
\begin{figure*}[p]
  \begin{minipage}{.24\textwidth}
    \centering
    \includegraphics[width=\linewidth]{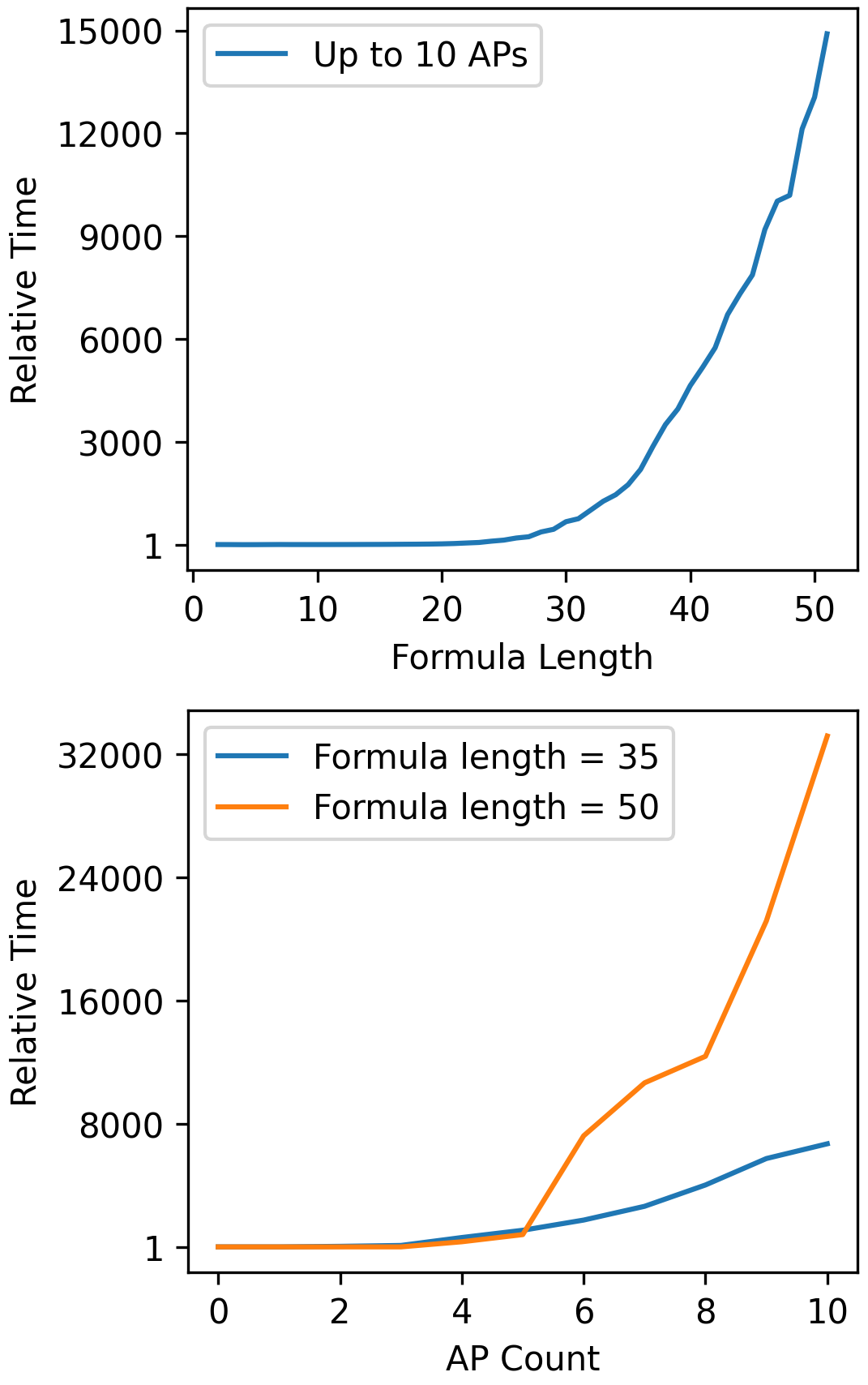}
    \caption{Scaling behavior of the trace generation using \texttt{spot}. \label{fig:tracegen}   }
  \end{minipage}
  \hfill
  \begin{minipage}{.75\textwidth}
    \centering
    \subfigure[LTL Solving]{\label{fig:ltl-abl-heatmap}\includegraphics[width=\textwidth]{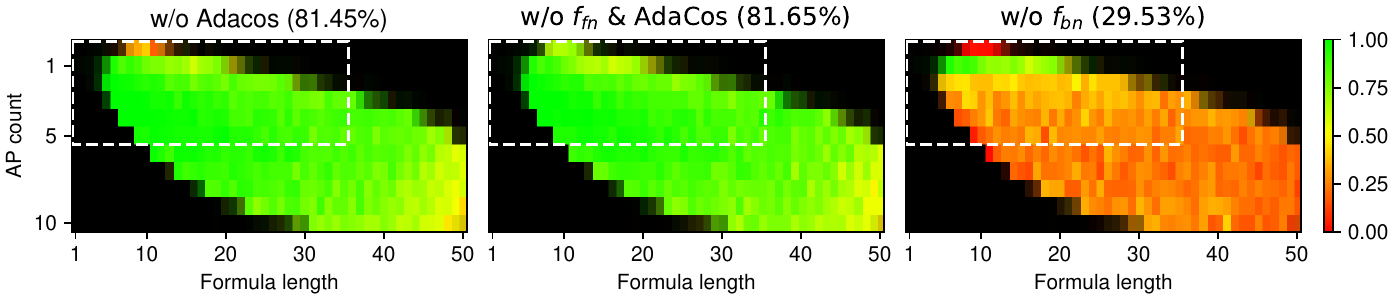}}
    \subfigure[Propositional Logic]{\label{fig:prop-abl-heatmap}\includegraphics[width=\textwidth]{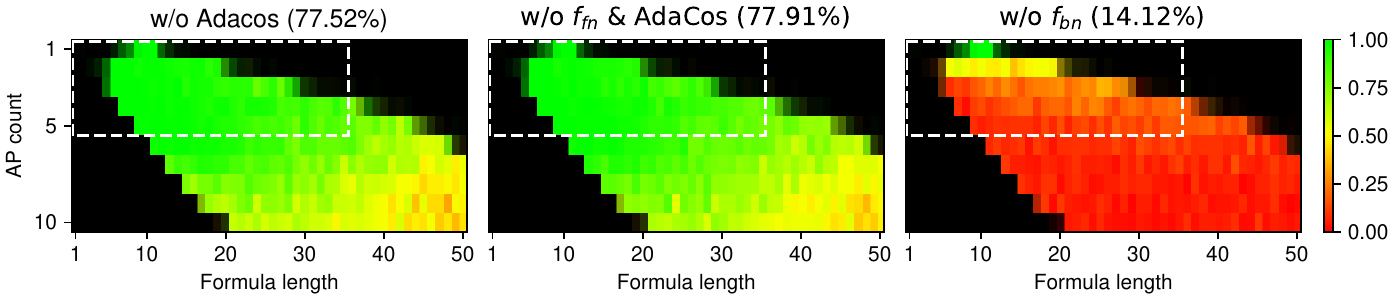}}
    \caption{
      Heatmaps for the ablation studies.
      The results are reported on the same test set as in Figure~\ref{fig:heatmaps}.
    }
  \label{fig:abl-heatmaps}
  \end{minipage}
\end{figure*}
\begin{figure*}[p]
  \begin{minipage}{.6\textwidth}
    \centering
    \includegraphics[width=\textwidth]{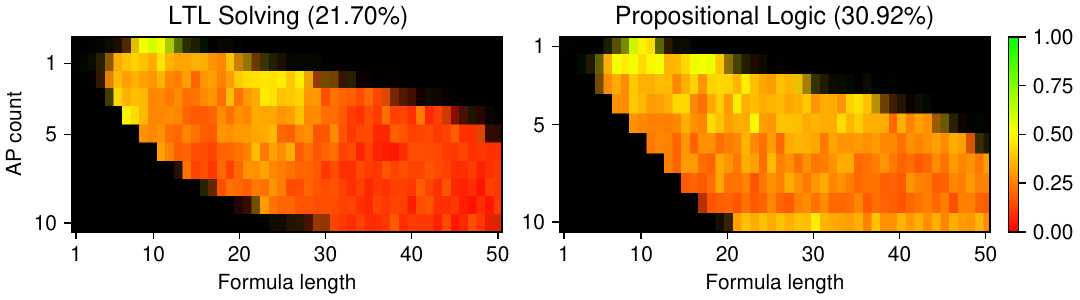}
    \caption{Llama 3.2 heatmaps for the two logic tasks. \label{fig:llama}   }
  \end{minipage}
  \hfill
  \begin{minipage}{.39\textwidth}
    \centering
    \includegraphics[width=.7\textwidth]{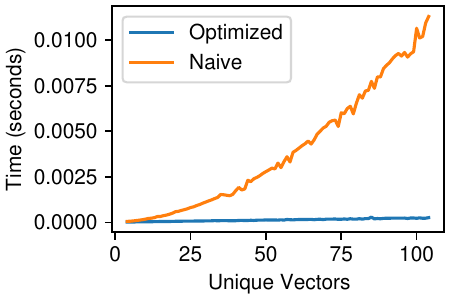}
    \caption{Average runtime cost of generating 8-dimensional unique random vectors from Neighboring Points with different uniqueness checking methods. \label{fig:runtime}   }
  \end{minipage}
\end{figure*}

\subsubsection{Alpha-Covariance}

Given a vocabulary of $n$ AP tokens and an LTL formula-trace pair containing $k$ APs, it's possible to write $\Perm{n}{k} = n! / (n-k)!$ alpha-equivalent pairs. %
Since these are semantically equivalent, we expect the model's predictions to be the same after undoing the alpha-conversions for all of them.
As there is no metric to quantify this in the literature to the best of our knowledge, we develop a new metric. %

Let $(\vx, \vy)$ be an input-output pair for the model, and let $\sP = \{ (\vx^1,\vy^1), \ldots, (\vx^n,\vy^n) \}$ be $n$ input-output pairs alpha-equivalent to $(\vx, \vy)$.
We define $\alpha_i$ as the alpha-conversion function for the $i$th input-output pair such that $\alpha_i(\vx) = \vx^i$ and $\alpha_i(\vy) = \vy^i$.
To compute the alpha-covariance of a model with respect to $\sP$, we generate predictions for each input in $\sP$,
obtaining the prediction $\hat{\vy}^i$ for each $x^i$.
We define a set that contains the predictions with alpha-conversion undone: $\sU = \{ \alpha_i^{-1}(\hat{\vy}^i) \mid 1 \leq i \leq n \}$.
Note that if we defined this set for the ground truth outputs in $\sP$, we would get $\{ \vy \}$ since $\alpha_i^{-1}(\vy^i) = \vy$ holds for each $\vy^i$ by definition.
The model's sensitivity to alpha-conversions could be quantified by simply $|\sU|$, but this value may be hard to interpret since it depends on $|\sP|$.
To normalize this value intuitively, we define the alpha-covariance of a model with respect to $\sP$ as in \Eqref{eq:alpha-covariance}.
\begin{equation}
  \label{eq:alpha-covariance}
  1 - \frac{|\sU| - 1}{|\sP| - 1}
\end{equation}

Intuitively, when alpha-covariance is $1$, the model is unaffected by all alpha-conversions in $\sP$.
An alpha-covariance of $0$ indicates that $|\sU| = |\sP|$, i.e., the model's prediction for each alpha-equivalent pair is unique after undoing the alpha-conversion.
This is unwanted because alpha-conversions should not change the semantic meaning.
Thanks to the embedding randomization in our method, an alpha-conversion does not necessarily change the embeddings, and conversely, there are multiple ways to embed the same input.%

For the proposed method, we generate the random embeddings once at the start of an evaluation run using the heuristic explained in Appendix \ref{sec:copy-randomness-sensitivity}.
Thus, alpha-conversions in this context are equivalent to shuffling the random embeddings in our method, which amounts to measuring our model's robustness against differences in random embeddings.

We report the results in Table \ref{perturbation-table}, which demonstrates that our method has a positive impact on the alpha-covariance, especially in limited data settings.
Since the LTLRandom35 dataset was created synthetically, it doesn't have any noteworthy biases and even the baseline enjoys a high alpha-covariance thanks to this.
However, when the dataset is perturbed by introducing a bias to the order of APs, the baseline struggles heavily with alpha-covariance, whereas our method does not.

\begin{table*}[htbp]
  \centering
  \caption{
    Mean alpha-covariance values for varying AP counts, evaluated on 1000 test samples, each with 120 random alpha-equivalent variants.
    The best value for each AP count is highlighted in bold.
  }
  \label{alpha-cov-10ap}
  \vskip 0.1in  %
  \scalebox{0.92}{
    \begin{tabular}{c|l|cccccccc}
        \multirow{2}*{\textbf{Task}} & \multicolumn{1}{c|}{\multirow{2}{*}{\textbf{Model}}} &  \multicolumn{8}{c}{\textbf{Alpha-Covariance}}\tabularnewline
         & & \textbf{3 AP} & \textbf{4 AP} & \textbf{5 AP} & \textbf{6 AP} & \textbf{7 AP} & \textbf{8 AP} & \textbf{9 AP} & \textbf{10 AP} \tabularnewline
    \hline 
        & Full Vocabulary & 54.09\% & 45.51\% & 45.23\% & \textbf{42.07}\% & 33.54\% & 34.47\% & 32.36\% & \textbf{28.42}\% \tabularnewline
        LTL & Alpha-Renaming & 50.64\% & 43.00\% & 40.95\% & 37.49\% & 30.80\% & 30.30\% & 28.76\% & 25.57\% \tabularnewline
        & Proposed & \textbf{54.30}\% & \textbf{46.05}\% & \textbf{45.64}\% & 41.88\% & \textbf{33.89}\% & \textbf{35.29}\% & \textbf{33.18}\% & 28.34\% \tabularnewline
    \hline 
        \multirow{3}*{\makecell{Propositional \\ Logic}} & Full Vocabulary & 39.77\% & 30.08\% & 30.37\% & 26.64\% & 20.97\% & 22.97\% & 18.80\% & 17.20\% \tabularnewline
        & Alpha-Renaming & 42.29\% & 32.36\% & 33.45\% & \textbf{30.28}\% & 24.91\% & 26.47\% & \textbf{22.29}\% & 19.83\% \tabularnewline
        & Proposed & \textbf{43.36}\% & \textbf{32.49}\% & \textbf{33.65}\% & 30.04\% & \textbf{25.00}\% & \textbf{26.63}\% & 21.99\% & \textbf{20.75}\% \tabularnewline
    \end{tabular}
  }
\end{table*}

\subsubsection{Generalization}
\label{sec:experiments:ltl:generalization}

The test dataset for this experiment contains at most 100 formula-trace pairs for each combination of AP count and formula length, whose maximum is 50 instead of 35.
We report the results for our model (using Hypercube Vertices, $d_\beta = 5$) and the three baselines in Figure \ref{fig:ltl-heatmap}.
The first baseline uses the same training dataset, whereas the second baseline uses a new LTL dataset with 10 APs, which we create using the same method as LTLRandom35.
For the third baseline, we train a fixed embedding model with 10 APs using the same 5 AP dataset but we shuffle the AP embeddings in each forward pass during training.

\textbf{Discussion.}
Despite seeing only 5 APs during training, our method performs only slightly worse than the full vocabulary baseline, which represents what a transformer-based model can do with 10 APs.
Our method outperforms both the vanilla and the alpha-renaming baselines by a considerable margin, which is significant since the latter is the only other model that can generalize to more APs.
Based on this, we hypothesize that the proposed stochastic AP embeddings provide a more explicit enforcement towards learning embedding-covariant transformations in the model, as opposed to training with alpha-renaming, where the learned embeddings may still carry unwanted token-specific biases.
Furthermore, unlike the baseline models, our model does not have to learn the concept of AP from scratch for each AP token thanks to the shared embedding part.
This could explain why our method shone against the alpha-renaming baseline in the LTL task where the interchangeable tokens are more complex than the copying task.

\textbf{Motivation for generalization.}
The generalization to larger AP counts is important especially when considering the exponential growth of the dataset generation time.
In Figure \ref{fig:tracegen}, we visualize the growth pattern of the trace checking duration based on increasing formula length and AP count.
The times are relative to the fastest trace checking time.
The exact times will vary depending on the machine.
In our experiments, generating 100000 samples of exact formula length 50 with at most 10 APs took 2 hours and 21 minutes on a system with 56 threads.

\textbf{Alpha-covariance.}
On the same generalization dataset,
we evaluate the alpha-covariance performance of these models in Table~\ref{alpha-cov-10ap}.
Note that since 10 APs lead to a lot more naming permutations than 5 APs, the alpha-covariance values are remarkably smaller compared to Table \ref{perturbation-table}.
Unlike the results from Table \ref{perturbation-table}, however, our method outperforms the alpha-renaming approach here.
This shows that our method excels in out-of-distribution settings, but trades off some in-distribution performance.
Although the full vocabulary baseline performs very similarly to our method, it's important to note that this region is in-distribution for that model.
Overall, these results align with Figure \ref{fig:ltl-heatmap}.

\subsection{Assignment Prediction for Propositional Logic}
\label{sec:experiments:prop}

To further demonstrate the applicability and generalization capabilities of our method, we evaluate it on a considerably different logical problem: predicting assignments for propositional logic (Appendix \ref{app:prop}).
The experimental setup is based on DeepLTL \citep{deepltl} with minor differences in hyperparameter choices (Appendix \ref{app:hyperparam}).
We use \verb|pyaiger| \citep{pyaiger} to generate datasets and evaluate predictions.
In Appendix \ref{app:propexp:setup}, we provide additional details about our experimental setup.

We perform the generalization experiment as in Section \ref{sec:experiments:ltl:generalization} and report the results in Figure \ref{fig:prop-heatmap}.
We observe the same ranking with slightly larger performance gaps.
Once more, the proposed method is superior to all approaches that use the same 5 AP training dataset, beaten only by the full vocabulary model which sidesteps the challenge of AP generalization due to its enhanced training dataset.

We continue propositional logic experiments in Table \ref{alpha-cov-10ap} and Appendix \ref{app:propexp:perturbation}, which focus on alpha-covariance and dataset perturbations respectively.
The results of these experiments also align with the LTL experiments.

\subsection{Ablation Studies}
\label{sec:experiments:ablation}

The hyperparameter search in Appendix~\ref{app:copying:hyperparam-search} operates on the copying task, and, alongside searching over the embedding hyperparameters, experiments with disabling the normalization features and AdaCos, thereby constituting an ablation study.
For the LTL and propositional logic tasks, we always kept the normalization features and AdaCos enabled in the previous sections.
In this section, we evaluate the impact of these features by disabling them on our best-performing models for these two logic tasks.
We ablate one aspect at a time, except for $f_{fn}$, which is disabled together with AdaCos because AdaCos depends on $f_{fn}$ to function correctly.

Figure \ref{fig:abl-heatmaps} presents the results, which demonstrate the critical importance of the $f_{bn}$ normalization component.
Removing $f_{bn}$ leads to dramatic performance drops (from 90.76\% to 29.53\% on LTL, and from 77.70\% to 14.12\% on propositional logic), confirming that maintaining balance between the common and randomized embedding parts is essential for our method's success.
The experiments with AdaCos and $f_{fn}$ indicate task-dependent benefits: they provide significant improvements on LTL (90.76\% vs. 81.45\% when AdaCos is removed), while showing negligible impact on propositional logic.

\subsection{Comparison with LLMs}

To contextualize the effectiveness of our proposed approach, we evaluate the performance of a general-purpose LLM (large language model), specifically, the 3B parameter version of Llama 3.2 \citep{llama3herd}, on the LTL task.
The details about the prompt design, inference parameters, and implementation are provided in Appendix~\ref{app:llama}.

In the last row of Table \ref{perturbation-table}, we report the performance of Llama 3.2 on the test split of LTLRandom35.
These results (e.g., 24.33\% correct) are drastically lower than those achieved by our proposed method (95.94\%).
On propositional logic, Llama 3.2 achieves a slightly better accuracy but much worse alpha-covariance (Table \ref{prop-perturbation-table} in Appendix \ref{app:propexp:perturbation}).
Additionally, we replicate the setups in Figure~\ref{fig:heatmaps} using Llama 3.2 on the same datasets and sample sizes. As shown in Figure~\ref{fig:llama}, the resulting accuracies are 21.70\% (LTL solving) and 30.92\% (propositional logic), compared to 90.76\% and 77.70\% by our method. This striking gap illustrates the limitations of general-purpose LLMs in highly specialized domains such as LTL solving, even when the model size far exceeds that of our dedicated architectures.

\subsection{Computational Efficiency}
\label{sec:experiments:efficiency}

We evaluate the computational efficiency of our method in terms of training time, inference speed, and memory usage (see Appendix~\ref{app:efficiency} for full details).
Our method incurs a modest 13\% training overhead compared to the baseline in LTL solving task.
At inference, embedding preparation takes only 0.0003 seconds and is required just once at the beginning of an evaluation session, making its cost negligible relative to model execution (0.206 seconds for a forward pass and 9.808 seconds for autoregressive generation).
Our optimized method for generating unique random vectors with integer reservoir sampling (Section~\ref{sec:method:random}) scales efficiently to a large number of vectors unlike the naive approach (Figure~\ref{fig:runtime}).
While the parameter count of traditional embeddings scales linearly with interchangeable token count, our method's parameter count remains constant, as embeddings are shared across interchangeable tokens.

%% file: sections/table_alpha_cov.tex
\begin{table*}[htbp]
  \def\arraystretch{1.1}
  \caption{
    Evaluation of the baselines, our method, and Llama 3.2 on the LTLRandom35 dataset.
    The alpha-renaming baseline was trained using 5 AP embeddings since vocabulary generalization is not evaluated here.
    First two columns denote the training dataset and the model.
    Next two columns indicate the ratio of the correct predictions and exact matches on 99,989 test set samples as evaluated by \texttt{spot}.
    Last three columns display mean alpha-covariance values for varying atomic proposition (AP) counts, evaluated on all alpha-equivalent variants of 1000 test samples.
    The results indicate that our method induces a robust inductive bias for alpha-equivalence.
  }
  \label{perturbation-table}
  \vskip 0.1in  %
  \begin{center}
    \begin{tabular}{ll|rr|rrr}
    \textbf{Training} &  & \multicolumn{2}{c|}{\textbf{Evaluation}} & \multicolumn{3}{c}{\textbf{Alpha-Covariance}}\tabularnewline
\textbf{Dataset} & \textbf{Model} & \textbf{Correct} & \textbf{Exact} & \textbf{3 AP} & \textbf{4 AP} & \textbf{5 AP}\tabularnewline
    \hline 
    Normal    & Baseline &         98.23\%  &         83.23\%       &         96.87\%  &         95.86\%  &         91.80\% \tabularnewline
    \hline 
    Perturbed & Baseline &         34.13\%  &         12.12\%       &         64.93\%  &         57.99\%  &         40.91\% \tabularnewline
Perturbed & Alpha-Renaming & \textbf{97.96\%} & \textbf{77.66\%} & \textbf{99.55\%} & \textbf{99.49\%} & \textbf{98.86\%}\tabularnewline
    Perturbed & Proposed &         95.94\%  &         76.45\%       &         97.66\%  &         97.76\%  &         98.29\% \tabularnewline
    \hline 
    Pretrained    & Llama 3.2 3B &         24.33\%  &         0.34\%       &         68.17\%  &         63.27\%  &         62.34\% \tabularnewline
    \end{tabular}

  \end{center}
\end{table*}

%% file: sections/limitations.tex
\section{Limitations}
\label{sec:limitations}

While our method provides an effective framework for enforcing alpha-equivalence in formal languages, it is not directly applicable to natural language, in which tokens carry semantic and contextual information that is often essential for interpretation.
For instance, even though variable names like \texttt{electricity\_bill} and \texttt{water\_bill} may be functionally interchangeable in certain code constructs, they convey distinct meanings that are not preserved under alpha-conversions. %
As such, enforcing alpha-equivalence may reduce interpretability and degrade performance in tasks that rely on linguistic connotations.
This represents an intriguing area for future research.

Another limitation is the requirement to manually define the set of interchangeable tokens, which may not be feasible in some settings. %
Moreover, our method requires training from scratch due to modifications in the embedding architecture, posing challenges for integration with pretrained models.

Although our dual-part embedding method demonstrates generalization capabilities, its performance in the LTL solving task decreases slightly for in-distribution data (Table~\ref{perturbation-table}).
The future work can tackle this issue, which may eventually lead to Pareto improvements in bias-variance tradeoff.
Finally, new randomization and normalization methods for our embeddings can be explored.

%% file: sections/conclusion.tex
\section{Conclusion}
\label{sec:conclusion}

A central goal in machine learning is to generalize to out-of-distribution samples, for which the model design and its inductive biases play a vital role.
In this work, we tackle the challenge of generalizing to larger vocabulary sizes unseen during training and creating an inductive bias for alpha-equivalence.
We also contribute the alpha-covariance metric for measuring the model consistency against alpha-equivalent inputs.
These contributions embody a foundation for learning extensible vocabularies for interchangeable tokens, which is especially useful for formal reasoning tasks in which alpha-equivalence naturally arises.

%% file: sections/impact.tex
\section*{Impact Statement}

This paper presents work whose goal is to advance the field of Machine Learning. There are many potential societal consequences of our work, none which we feel must be specifically highlighted here.

%% file: sections/prelim.tex
\section{Preliminary: Language models}
\label{app:llm-prelim}

The autoregressive language modeling or sequence modeling in a broader sense—whose goal is to predict the next token given the past tokens—was revolutionized by the transformer architecture~\citep{vaswani}, replacing the step-by-step processing of recurrent neural networks (RNNs) with a parallelizable attention mechanism.
At its core lies the attention mechanism, which computes three vectors—query, key, and value—from input embeddings.
This mechanism allows the model to weigh the importance of different tokens, enabling it to capture long-range dependencies efficiently.
In self-attention, these vectors come from the same sequence, while in cross-attention, key and value vectors come from a different sequence, as in encoder-decoder setups.
The transformer consists of an encoder with self-attention and feed-forward layers, and a decoder that adds cross-attention to incorporate the encoder’s output.
Since attention lacks an inherent sense of token order, positional encodings are added to input embeddings to provide sequence structure.
During training, attention masking ensures causality in predictions, preventing future tokens from being considered when predicting the next one.

%% file: sections/ltl.tex
\section{Temporal logic overview}
\label{app:ltl}

Linear Temporal Logic (LTL) extends propositional logic by introducing the ability to reason about the evolution of propositions over time~\citep{Pnueli77}.
The syntax of LTL, defined over a finite set of atomic propositions $P$, is given in \Eqref{eq:ltl},
where $\mathbf{T}$ represents \textit{True}, $p \in P$ an atomic proposition, $\neg$ the negation operator, $\wedge$ the conjunction operator,
$\X$ and $\until$ the temporal operators \textit{next} and \textit{until} respectively.

\begin{equation}
  \label{eq:ltl}
  \phi := \mathbf{T} \mid p \mid \neg \phi \mid \phi_1 \wedge \phi_2 \mid \X \phi \mid \phi_1 \until \phi_2
\end{equation}

Specifically:
\begin{itemize}
  \item $\X \phi$ holds at time $t$ if and only if $\phi$ holds at the next time step, i.e., at time $t+1$.
  \item $\phi_1 \until \phi_2$ means that $\phi_2$ must hold at some future time $t_2$, and $\phi_1$ holds at every time step $t$ from the current time $t_1$ up to but not necessarily including $t_2$.
\end{itemize}

For instance, the formula $\X \X a$ specifies that $a$ must hold at the third time step. Similarly, the formula $\mathbf{T} \until a$ requires that $a$ holds at some point in the future.
Finally, as a more complex example, the formula $\X b \wedge a \until c$ asserts that $b$ holds at the second time step, $c$ holds at some future time, and $a$ holds at all preceding time steps.

An LTL formula is evaluated over a \textit{trace}, which represents a sequence of truth values for atomic propositions over time.
In this work, as in DeepLTL~\citep{deepltl}, we consider \textit{symbolic} traces of \textit{infinite} length.
These traces are expressed in what is known as a \textit{lasso} form, denoted $uv^\omega$, where $u$ is a finite prefix, and $v$ is a finite sequence that repeats indefinitely.

A symbolic trace represents all traces that satisfy the propositional formulae at the respective time steps. For example, the symbolic trace $a, a \wedge \neg b, (c)^\omega$ describes all traces in which $a$ holds at the first two time steps, $b$ does not hold at the second time step, and $c$ holds at every step from the third onward.
This symbolic trace satisfies the formulae $\mathbf{T} \until c$ and $\X \neg b \wedge a \until c$, but it violates the formula $\X \X b$ since $b$ is not guaranteed to hold at the third time step.
Symbolic traces, such as this one, can be underspecified, meaning that certain propositions (e.g., $a$ and $b$) may take arbitrary values at some time steps.

The LTL solving problem involves identifying a symbolic trace in lasso form $uv^\omega$ that satisfies a given input formula $\phi$.
We approach this as an autoregressive language modeling task: given an LTL formula and a partially generated symbolic trace, the model predicts the probabilities for the next token in the trace.

For compatibility with the dataset from DeepLTL~\citep{deepltl}, both our traces and formulae are represented in Polish (prefix) notation, where operators precede their operands. For instance, $a \wedge b$ is written as \verb|&ab|, which avoids the need for parentheses to resolve ambiguities.

As described earlier, we assume that traces are infinite and represented in lasso form $uv^\omega$. Alongside atomic propositions, constants (\texttt{True:1} and \texttt{False:0}), and logical operators, we use special symbols in the notation: ``\verb|;|'' is a position delimiter, and ``\verb|{|'' and ``\verb|}|'' enclose the repeating period $v$. For example, the string ``\verb|a;&ab;{b}|'' represents the symbolic trace $a, a \wedge b, (b)^\omega$.

%% file: sections/prop.tex
\section{Propositional Logic}
\label{app:prop}

Unlike LTL (Appendix \ref{app:ltl}), propositional logic does not feature any temporal operators, but we include the derived operators for equivalence ($\leftrightarrow$) and exclusive or ($\oplus$) alongside the basic negation ($\neg$), conjunction ($\wedge$), and disjunction ($\vee$).
This leads to the syntax given in \Eqref{eq:prop}, defined over a finite set of atomic propositions $P$ where $p \in P$ an atomic proposition.

\begin{equation}
  \label{eq:prop}
  \phi := \mathbf{T}
  \mid p
  \mid \neg \phi
  \mid \phi_1 \wedge \phi_2
  \mid \phi_1 \vee \phi_2
  \mid \phi_1 \leftrightarrow \phi_2
  \mid \phi_1 \oplus \phi_2
\end{equation}

In assignment prediction problem for propositional logic, the goal is to determine a Boolean assignment for every atomic proposition $p \in P$ such that the given formula is satisfied.
We allow the assignments to be partial, e.g., just as $a = 1$, $b = 1$ is a valid assignment for the formula $a \vee b$, so is $a = 1$, which allows $b$ to take any value.

To encode the assignments for the neural network, an alternating sequence of atomic propositions and values is used.
For example, \verb|a1b0| represents the assignment $a = 1$ and $b = 0$.
To verify the outputs of the neural network and to generate datasets, \verb|pyaiger| was used \citep{pyaiger}.

%% file: sections/hyperparam.tex
\section{Hyperparameters}
\label{app:hyperparam}

The constant hyperparameter choices for all experiments are given in Table~\ref{hyperparam-table}.
These hyperparameters are kept constant within an experiment.
The hyperparameters for the logic tasks are taken from DeepLTL \citep{deepltl}.
For the LTL task, we used the same hyperparameters.
On the other hand, for the propositional logic task, we had to make some changes to adapt them to our updated architecture.
Firstly, since we utilize weight sharing, we cannot separate the embedding dimensions of encoder and decoder.
As a result, instead of having an embedding dimension of 128 for the encoder and 64 for the decoder, we use 128 for both.
However, since there are 6 attention heads, we round it up to 132.

\begin{table}[htbp]
  \centering
  \caption{Hyperparameter choices.}
  \label{hyperparam-table}
  \vskip 0.1in  %
\begin{tabular}{c | c c c  c | c c }
  Experiment & Embedding & Layers & Heads & FC size & Batch Size & Train Steps \\
  \hline
  Copy (Sections \ref{sec:copy-1} and \ref{sec:copy-2}) & 64        & 2      & 4     & 64      & 512        & 20K \\
  Copy (Section~\ref{sec:copy-big}) & 128       & 6      & 8     & 128     & 512        & 20K \\
  LTL  (Section~\ref{sec:experiments:ltl}) & 128       & 8      & 8     & 1024    & 768        & 52K \\
  Propositional Logic (Section~\ref{sec:experiments:prop}) & 132       & 6      & 6     & 512    & 1024        & 50K \\
\end{tabular}
\end{table}

%% file: sections/copying.tex
\section{Copying Task Experiments}
\label{app:copying}

\textbf{Evaluation method.}
We generate the predictions using greedy sampling in the copying task.
We use the edit distance between the prediction and the ground truth as our evaluation metric.
To generate the evaluation datasets (validation and test splits), we create 100 samples for each possible combination of unique character count and string length, starting from a minimum of 3.
Consequently, the total evaluation dataset is arranged in a matrix in which the rows represent unique character count in the string and the columns represent the string length.
This matrix is upper triangular since the unique character count cannot exceed the string length.
For random embeddings, we repeat the evaluation 10 times and report the average.
To evaluate up to the string length of 30 in this setup, $10 \times 100 \times 406 = 406000$ predictions are required, where $406$ is the number of upper triangular elements in a $28 \times 28$ matrix.
To minimize the impact of random factors, we train each model three times and report the results only for the best.

\subsection{Generalization to larger vocabularies}
\label{sec:copy-1}
We create a dataset consisting of 10 million strings whose lengths vary between 3 and 30 with at most 5 unique characters.
We evaluate the models on strings up to length 30 with at most 30 unique characters.
Out of 27 models we trained with dual-part embeddings, 20 of them achieve an average edit distance of 0.0, i.e., no error.
The worst model's average edit distance is 1.0.
For comparison, an output sequence of length 30 can have a maximum edit distance of 30.

\subsection{Generalization to larger vocabularies and lengths}
\label{sec:copy-2}
We create a dataset consisting of 10 million strings whose lengths vary between 5 and 10 with at most 5 unique characters.
We evaluate on the same validation set as before, expecting the model to generalize to both longer lengths and larger vocabulary sizes.
In the next subsection, we perform a hyperparameter search over random embedding methods, $d_\beta$ values, and whether $f_{bn}$, $f_{fn}$, AdaCos are enabled.

\subsection{Hyperparameter Search}
\label{app:copying:hyperparam-search}

On the smaller copying task,
we train multiple models that use different random embedding methods (Section~\ref{sec:method:random}) with different $d_\beta$ values.
While altering $d_\beta$, we keep the total number of embedding dimensions $d_\alpha + d_\beta$ constant.
We train each model at least 3 times with different seeds and report the results for the best one in Tables \ref{hyperparam-hell} (proposed method) and \ref{baseline-hyperparam-hell} (baselines).

\begin{table}[htbp]
\begin{center}
  \caption{
    Mean edit distance for various models using proposed method.
    The numbers in the header row represents $d_\beta$ for each random embedding method.
    In the first column, enabled normalization features are listed.
    AC refers to AdaCos, which can only be enabled when $f_{fn}$ is used.
  }
  \label{hyperparam-hell}
  \vskip 0.1in  %

\scalebox{0.9}{
\begin{tabular}{c|ccccc|ccccc|ccccc}
  Enabled & \multicolumn{5}{c|}{Normal Distribution} & \multicolumn{5}{c|}{Neighboring Points} & \multicolumn{5}{c}{Hypercube Vertices} \\
  Features        & 2 & 4 & 8 & 16 & 32 & 4 & 6 & 8 & 16 & 32 & 5 & 6 & 8 & 16 & 32 \\
    \hline
    $f_{bn}$ + $f_{fn}$ + AC     & 13.6   & 5.4    & 4.6    & 8.1    & 8.1    & 1.9    & 13.0   & 2.2    & 1.0    & 2.1    & 2.8    & 0.4    & 7.5    & 8.4    & 3.9   \\
    $f_{fn}$ + AC                & 7.6    & 13.1   & 4.6    & 2.2    & 5.2    & 8.7    & 11.5   & 2.8    & 2.9    & 2.2    & 0.5    & 3.7    & 3.2    & 4.2    & 4.1   \\
    $f_{bn}$ + $f_{fn}$          & 13.7   & 10.6   & 8.3    & 3.8    & 11.8   & 11.9   & 5.7    & 3.7    & 7.4    & 8.3    & 2.2    & 13.1   & 21.5   & 19.4   & 20.9  \\
    $f_{fn}$                     & 15.4   & 10.6   & 8.2    & 3.7    & 10.1   & 8.1    & 12.3   & 6.4    & 13.4   & 9.9    & 2.5    & 1.7    & 12.5   & 2.1    & 12.8  \\
    $f_{bn}$                     & 10.6   & 16.6   & 11.8   & 6.9    & 8.2    & 5.8    & 3.0    & 0.6    & 7.8    & 14.3   & 12.8   & 13.8   & 19.4   & 22.9   & 11.6  \\
    -                            & 16.5   & 11.6   & 12.6   & 12.5   & 9.0    & 12.5   & 3.7    & 9.5    & 5.9    & 13.5   & 12.7   & 9.6    & 8.6    & 15.9   & 16.6  \\
\end{tabular}
  }
\end{center}
\end{table}

\begin{table}[htbp]
\begin{center}
  \caption{
    Mean edit distance for various baseline models.
    In the first column, enabled normalization features are listed.
    AC refers to AdaCos, which can only be enabled when $f_{fn}$ is used.
    Note that $f_{bn}$ is not applicable for baseline models.
    The results for the first type of baseline are omitted since it cannot generalize to larger vocabularies.
    The second baseline was trained on a dataset with a vocabulary size of 30.
    The third baseline uses the same limited vocabulary dataset like the proposed method, but uses alpha-renaming as data augmentation.
  }
  \label{baseline-hyperparam-hell}
  \vskip 0.1in  %

\begin{tabular}{c|c|c}
    Enabled        & Baseline & Baseline \\
    Features       & 2nd Type & 3rd Type \\
    \hline
    $f_{fn}$ + AC  & 6.1 & 1.9 \\
    $f_{fn}$       & 4.9 & 11.3 \\
    -              & 5.5 & 12.9 \\
\end{tabular}
\end{center}
\end{table}

The results in Tables \ref{hyperparam-hell} and \ref{baseline-hyperparam-hell} exhibit high variance with no clear patterns that indicate which methods are better.
Therefore, we perform an analysis based on correlation coefficients between these hyperparameters and the edit distance using the results from all 277 models we've trained (not including the baseline models).
For this analysis, we assume that the value of Boolean properties (such as $f_{bn}$, $f_{fn}$ and AdaCos) are 0 or 1.
The correlation coefficients are as follows:
\begin{center}
\begin{tabular}{ccc|c|cc|c}
  N.D. & N.P. & H.V. & $d_\beta$ & $f_{bn}$ & $f_{fn}$ & AdaCos \\ %
  \hline
  0.02 & -0.14 & 0.11 & 0.01 & 0.10 & -0.29 & -0.41 \\ %
\end{tabular}
\end{center}
First three columns are the random embedding methods as listed in Table \ref{random-methods-table}, the fourth column is $d_\beta$, and the last three columns represent whether the given feature is enabled.
Accordingly, the best random embedding method is ``Neighboring Points'' since it's the only one that correlates negatively with edit distance.
The correlation observed for $d_\beta$ is negligible.
Introducing $f_{bn}$ increases the edit distance, but the statistical significance is not ideal (p-value 0.04).
Both $f_{fn}$ and AdaCos loss have a positive and statistically significant impact on edit distance, with p-values smaller than $10^{-6}$.

\begin{figure*}[htbp]
  \centering
  \includegraphics[width=\textwidth]{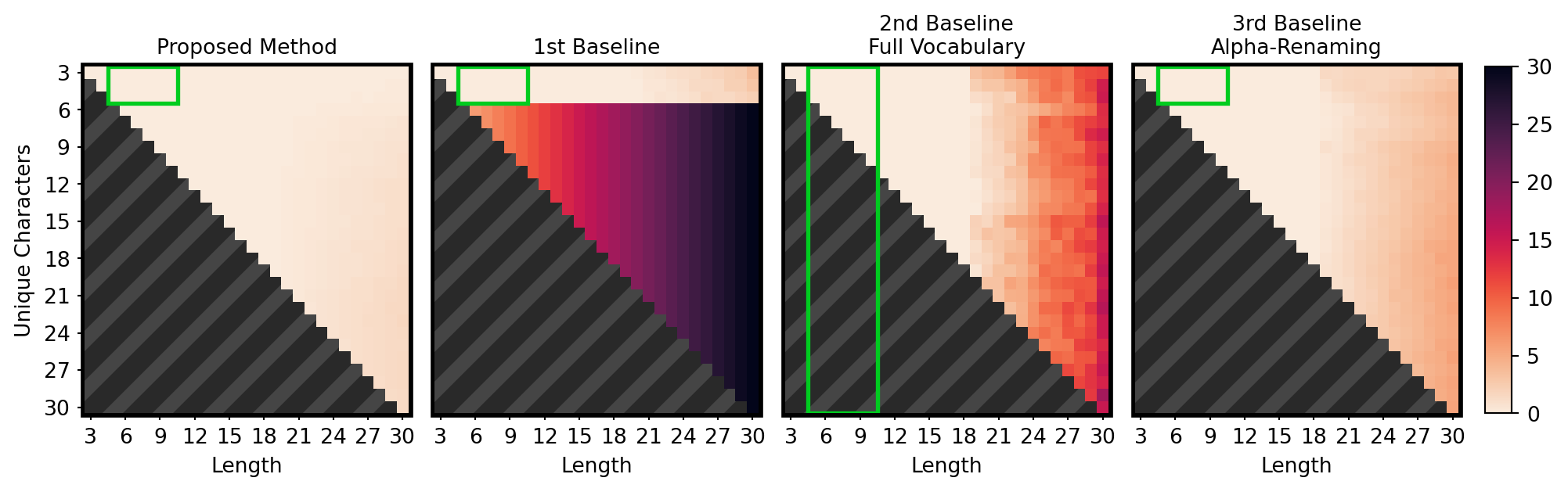}
  \caption{
    Edit distance heatmaps on test set.
    The first and second heatmaps are the proposed and baseline (first type) models respectively, trained on strings up to length 10 and a vocabulary size 5.
    The third heatmap is the second baseline, which uses a new training dataset with a larger vocabulary.
    The last heatmap is the third baseline that uses the same dataset as the proposed method but incorporates alpha-renaming in training.
    The difference between the last two baselines is that the alpha-renaming baseline is not exposed to more than 5 unique characters per sample.
    The lower triangular part of each heatmap (gray hatch pattern) represents the impossible combinations of length and unique character count.
    The green box represents the number of unique characters (y-axis) and the maximum length (x-axis) in the training dataset.
    Note that all heatmaps share the same y-axis.
  }
  \label{fig:cpy-triple}
\end{figure*}

We determine the best model for the proposed method and the baseline on the validation set, evaluate them on the test set and visualize the results in Figure \ref{fig:cpy-triple}.
Since the baseline model cannot process larger vocabularies, we assume that the prediction is empty if the unique character count exceeds the training set's vocabulary, hence the edit distance equals length in that area.
Our best model trained on limited length uses Hypercube Vertices with $d_\beta$ set to $6$ and $f_{fn}$ + AdaCos enabled.
It achieves a mean edit distance of 0.38 on the test set.
The first baseline's mean edit distance is 0.51 (calculated up to 5 unique characters, only for this model).
The second and third baselines' mean edit distances are 4.93 and 1.85 respectively.
However, the significance of this difference is highly questionable, as these models exhibit high variance across different training runs.

\subsection{Sensitivity to randomness in embeddings}
\label{sec:copy-randomness-sensitivity}

We analyze the impact of the randomization that the proposed method performs on embeddings.
The minimum, mean, and maximum edit distance (on test set) obtained by ten different embedding randomizations of the second model in Figure \ref{fig:cpy-triple} are 0.25, 0.38, 0.55 respectively, with a sample standard deviation of 0.09.
The pooled standard deviation of the edit distance across all 277 models evaluated on the validation set is 1.73.
However, our best models are more resilient against randomness: this value is 0.74 for top 10\% models.

To reduce the computational cost of evaluation in other experiments (All LTL experiments and Section \ref{sec:copy-big}), we generate 10 random embeddings, sort them by their cross entropy loss on the evaluated dataset, and use the median one.
We find that this serves as a decent proxy for the average performance.
Across the validation set evaluations of all 277 models, the percent difference in edit distance between this median method and the real mean is 1.4\% on average (meaning that the result from the median method is worse), and 9.1\% if we consider the absolute differences.

\subsection{Scaling up}
\label{sec:copy-big}

We increase the length of the strings from 5-10 to 20-80, and vocabulary size from 5 to 20.
We create the evaluation sets by generating 20 samples for each combination of unique character count and string length.
The mean edit distance of our best model is 0.0.
The heatmap is given in Figure \ref{fig:intro-result}.
All baselines also attain perfect performance in this task on the vocabulary sizes they support.
Therefore, only the first type of baseline is shown in Figure \ref{fig:intro-result}.

%% file: sections/limited.tex
\section{LTL Experiment with Limited Dataset}
\label{app:ltl-limited}

This is a continuation of the experiment from Section \ref{sec:perturbation}.
Table \ref{limited-table} contains evaluations of the baseline, the alpha-renaming model, and the proposed model trained with a severely limited number of samples: 80,000 instead of 799,909.
We kept the number of epochs constant, and as a result, the number of training steps were also divided by ten (approximately).

The result of limiting the number of training samples is similar to the dataset perturbation, albeit much less pronounced for the baseline model.
Unlike in the perturbation experiment, where the baseline model's performance plummets, all models trained on the reduced dataset maintain similar correctness ratios.
The biggest difference is observed in the alpha-covariance values, particularly in the 5 AP category, whose ranking aligns with the perturbation experiment.

Since LTLRandom35 is a synthetic dataset, it exhibits minimal inherent bias, even when the dataset size is limited.
Consequently, limiting the dataset size has a smaller effect than introducing perturbations.
Furthermore, since the alpha-renaming model was trained using 5 AP embeddings in this experiment, it loses its vocabulary generalization capability unlike our proposed method.
Training the alpha-renaming baseline with more APs would require learning a new embedding for each AP, which would reduce its performance.

\begin{table}[htbp]
  \def\arraystretch{1.1}
  \caption{
    Evaluation of the baselines and our method trained on different versions of LTLRandom35.
    The same results from Table \ref{perturbation-table} are shown for easier comparison.
    The alpha-renaming baseline was trained using 5 AP embeddings since vocabulary generalization is not evaluated here.
    First two columns denote the training dataset and the model.
    Next two columns indicate the ratio of the correct predictions and exact matches on 99,989 test set samples as evaluated by \texttt{spot}.
    Last three columns display mean alpha-covariance values for varying atomic proposition (AP) counts, evaluated on all alpha-equivalent variants of 1000 test samples.
  }
  \label{limited-table}
  \vskip 0.1in  %
  \begin{center}
    \begin{tabular}{ll|cc|ccc}
    \textbf{Training} &  & \multicolumn{2}{c|}{\textbf{Evaluation}} & \multicolumn{3}{c}{\textbf{Alpha-Covariance}}\tabularnewline
\textbf{Dataset} & \textbf{Model} & \textbf{Correct} & \textbf{Exact} & \textbf{3 AP} & \textbf{4 AP} & \textbf{5 AP}\tabularnewline
    \hline 
    Normal    & Baseline &         98.23\%  &         83.23\%       &         96.87\%  &         95.86\%  &         91.80\% \tabularnewline
    \hline 
    Perturbed & Baseline &         34.13\%  &         12.12\%       &         64.93\%  &         57.99\%  &         40.91\% \tabularnewline
Perturbed & Alpha-Renaming & \textbf{97.96\%} & \textbf{77.66\%} & \textbf{99.55\%} & \textbf{99.49\%} & \textbf{98.86\%}\tabularnewline
    Perturbed & Proposed &         95.94\%  &         76.45\%       &         97.66\%  &         97.76\%  &         98.29\% \tabularnewline
    \hline 
    Limited   & Baseline &         87.47\%  &         63.61\%       &         94.37\%  &         91.70\%  &         85.64\% \tabularnewline
Limited & Alpha-Renaming & \textbf{89.50\%} & \textbf{64.15\%}      & \textbf{99.02\%} & \textbf{98.67\%} & \textbf{97.82\%}\tabularnewline
    Limited   & Proposed &         87.32\%  &         59.04\%       &         97.94\%  &         96.12\%  &         94.34\% \tabularnewline
    \end{tabular}

  \end{center}
\end{table}

%% file: sections/propexp.tex
\section{Propositional Logic Experiments}
\label{app:propexp}

This section gives more details about the experimental setup of the propositional logic task and continues the experiments.

\subsection{Experimental Setup}
\label{app:propexp:setup}

We use PropRandom35 from DeepLTL \citep{deepltl} as our main 5 AP dataset, and create other datasets using the same approach.
In particular, propositional logic formulae are generated randomly, with negation ($\neg$), conjunction ($\wedge$), and disjunction ($\vee$) operators having an equal weight.
Equivalence ($\leftrightarrow$) and exclusive or ($\oplus$) operators each have half as much weight since they are derived operators.
The corresponding assignment is generated by querying the \texttt{pyaiger}'s SAT solver for a minimal unsatisfiable core of the negated formula.

As in the LTL experiments, we use a transformer encoder-decoder architecture with three-way weight tying \citep{weight-tying}.
The positional encoding method is tree-positional encoding \citep{treepos} for the encoder and RoPE \citep{rope} for the decoder.
Predictions are generated using beam search with a beam size of 3.

Since the network outputs the assignments as a sequence (Appendix \ref{app:prop}),
the same assignment can be encoded in multiple ways by changing the order.
For example, both \verb|a1b0| and \verb|b0a1| represent the same set of assignments $a = 1$ and $b = 0$, which can be written as $\{ (a, 1), (b, 0) \}$ in set notation.
We consider such pairs exact matches in the propositional logic experiments.
If the predicted assignment does not exactly match the ground truth, we use \texttt{pyaiger} to evaluate the correctness.

\subsection{Dataset Perturbations}
\label{app:propexp:perturbation}

In this section, we repeat the dataset perturbation experiment (Section \ref{sec:perturbation}) for the propositional logic task.
The perturbation is introduced in a similar manner by renaming the APs such that the order of the first AP appearances in the label (sequence denoting the Boolean assignment) is always the same.
As shown in Table \ref{prop-perturbation-table}, the experimental results once again confirm that our method introduces a robust inductive bias for alpha-equivalence.

\begin{table*}[htbp]
  \def\arraystretch{1.1}
  \caption{
    Evaluation of the baselines, our method, and Llama 3.2 on the PropRandom35 dataset.
    The alpha-renaming baseline was trained using 5 AP embeddings since vocabulary generalization is not evaluated here.
    First two columns denote the training dataset and the model.
    Next two columns indicate the ratio of the correct predictions and exact matches on 100,000 test set samples as evaluated by \texttt{pyaiger}.
    Last three columns display mean alpha-covariance values for varying atomic proposition (AP) counts, evaluated on all alpha-equivalent variants of 1000 test samples.
  }
  \label{prop-perturbation-table}
  \vskip 0.1in  %
  \begin{center}
    \begin{tabular}{ll|rr|rrr}
    \textbf{Training} &  & \multicolumn{2}{c|}{\textbf{Evaluation}} & \multicolumn{3}{c}{\textbf{Alpha-Covariance}}\tabularnewline
\textbf{Dataset} & \textbf{Model} & \textbf{Correct} & \textbf{Exact} & \textbf{3 AP} & \textbf{4 AP} & \textbf{5 AP}\tabularnewline
    \hline 
    Normal    & Baseline   &         95.62\%  &         57.94\%       &        95.70\%  &      93.69\%  &     76.02\% \tabularnewline
    \hline 
    Perturbed & Baseline   &         41.57\%  &         9.04\%        &        14.96\%  &      16.85\%  &     10.65\% \tabularnewline
Perturbed & Alpha-Renaming &         93.85\%  &         57.24\%       &        99.56\%  &      99.60\%  &     93.23\% \tabularnewline
    Perturbed & Proposed   &         93.25\%  &         56.45\%       &        99.23\%  &      99.42\%  &     92.98\% \tabularnewline
    \hline 
    Pretrained    & Llama 3.2 3B &         29.03\%  &         1.56\%       &         50.75\%  &         27.96\%  &         11.25\% \tabularnewline
    \end{tabular}

  \end{center}
\end{table*}

%% file: sections/efficiency.tex
\section{Computational Efficiency Details}
\label{app:efficiency}

To evaluate the practical applicability of our method, we analyze its computational overhead compared to baseline approaches. We report training times, inference speeds, and memory requirements across different experimental settings.

\textbf{Training efficiency.}
We measured training durations for models trained on NVIDIA H100 GPUs using identical hyperparameter settings. In LTL solving task, the average training times were as follows:
\begin{itemize}
    \item Baseline (traditional embeddings): 2 hours 12 minutes
    \item Alpha-renaming baseline: 2 hours 33 minutes  
    \item Proposed method: 2 hours 29 minutes
\end{itemize}
The proposed method introduces minimal training overhead compared to the baseline, with only a 13\% increase in training time. This modest overhead stems from the additional embedding preparation steps required during training.

\textbf{Inference performance.}
We conducted a runtime analysis using our best-performing LTL model on NVIDIA A4000 hardware. The model uses Hypercube Vertices randomization with uniqueness checking enabled, evaluated with batch size 768 and beam search (beam size = 3).
In this setup, a forward pass takes 0.206 seconds, and autoregressive generation 9.808 seconds.
On the other hand, the embedding preparation time is measured at 0.0003 seconds, which is negligible compared to model execution.
Importantly, during inference, embeddings need only be generated once at the start of the evaluation session, making the amortized cost even smaller for batch processing.

\textbf{Memory overhead.}
Our method reduces the total parameter count compared to traditional approaches since only one common embedding is learned for all interchangeable tokens, regardless of their quantity.
The memory overhead comes primarily from constructing the embedding matrix during runtime, which requires temporary storage for the randomized components. However, this additional memory requirement is on the same order of magnitude as the embedding matrix itself, which represents a small fraction of total model parameters in transformer architectures.

The parameter efficiency of our method scales favorably with vocabulary size.
Unlike traditional approaches that require learning separate embeddings for each token (thereby scaling linearly with the vocabulary size), our method's parameter count remains constant regardless of the number of interchangeable tokens.
However, two factors require consideration for very large vocabularies:
\begin{enumerate}
    \item \textbf{Sampling set size}: In discrete random generation methods, the sampling set is naturally bounded (Table \ref{random-methods-table}). However, the sampling set grows exponentially with the number of dimensions, ensuring sufficient diversity even for large vocabularies.
    \item \textbf{Uniqueness checking}: For vocabularies with hundreds of thousands of tokens, uniqueness verification becomes computationally expensive, but the probability of collisions decreases exponentially with increasing embedding dimensions.
\end{enumerate}

%% file: sections/llama.tex
\section{LLM Setup}
\label{app:llama}

We use the 3B-parameter version of Llama 3.2 \citep{llama3herd}, quantized with \verb|Q4_K_M|, and run it using Ollama 0.4.7 as our LLM backend.
We first experimented with greedy sampling (by setting top-k=1) since Ollama does not support beam search.
However, we found that the default sampling options (top-k=40 and top-p=0.9) yielded better results.
Therefore, we use these default settings for all experiments.

Unlike our specialized models, which operate on prefix (Polish) notation, we prompt the LLM using infix notation for input formulas (and output traces in the LTL task), as this format is more prevalent in natural language and more familiar to general-purpose LLMs.
To output the assignments in the propositional logic task, we use JSON format, and constrain the LLM's output using a JSON  schema.
The input prompts for the LTL and propositional logic tasks are given in Listing~\ref{lst:llm-prompt-ltl} and Listing~\ref{lst:llm-prompt-prop}, respectively.
For each sample, the ``\verb|{formula}|'' substring in the prompt is replaced by the input formula, and the prompt is given as a user message to the LLM.

We set the random seed to 42 for each sample.
Although the reason behind this choice is reproducability, it also seems to improve alpha-covariance.
For example, the alpha-covariance values reported for Llama 3.2 in Table~\ref{perturbation-table} are
68.17\%, 63.27\%, 62.34\% for 3 to 5 APs, respectively, which decrease to
41.94\%, 43.10\%, 44.62\% when the random seed is no longer fixed.

\begin{lstlisting}[caption={LLM Prompt for the LTL solving task.},captionpos=t,label={lst:llm-prompt-ltl}]
Your task is to generate a satisfying trace for a given LTL (Linear Temporal Logic) formula.
Lowercase letters denote the atomic propositions.
The output trace should be in lasso form composed of two parts: the prefix part and the cycle part.
Timesteps in the trace should be separated by semicolons, and the cycle part should be enclosed in curly braces, preceeded by the keyword "cycle".

Temporal operators:
X: Next operator
U: Until operator

Logical operators:
&: AND operator
|: OR operator
!: NOT operator
The output trace is a symbolic trace, which means that the logical operators are allowed, but not temporal operators.

Constants:
0: False
1: True
Note that other numbers are invalid.

Example 1
Formula: X((a & Xa) U XXb)
Trace: 1; 1; 1; b; cycle{{1}}

Example 2
Formula: !c U X(1 U b)
Trace: 1; b; cycle{{1}}

Example 3
Formula: X!X!(b & Xb)
Trace: 1; 1; b; b; cycle{{1}}

Example 4
Formula: !(1 U !c)
Trace: cycle{{c}}

Your Turn
Formula: {formula}
Please generate the corresponding trace. Output the trace only.
\end{lstlisting}

\vspace{0.5em}

\begin{lstlisting}[caption={LLM Prompt for the propositional logic task.},captionpos=t,label={lst:llm-prompt-prop}]
Your task is to generate an assignment that satisfies a given propositional logic formula.
Lowercase letters denote the atomic propositions.
The output is a JSON object representing the assignment.

Logical operators (ordered from highest precedence to lowest):
!: NOT operator
&: AND operator
|: OR operator
xor: Exclusive OR operator
<->: Logical equivalence operator (biconditional)

Constants:
0: False
1: True
Note that other numbers are invalid.

Example 1
Formula: !a | c & (b <-> c)
Assignment: { "a": false }

Example 2
Formula: !(a <-> (!a xor !e))
Assignment: { "a": true, "e": true }

Example 3
Formula: a & (!a <-> !c | d)
Assignment: { "a": true, "c": true, "d": false }

Example 4
Formula: !(a | !(!d | b & d))
Assignment: { "a": false, "d": false }

Your Turn
Formula: {formula}
Please generate an assignment that satisfies this formula. Output the assignment only, in JSON format.
\end{lstlisting}

%% file: main.bbl
\begin{thebibliography}{33}
\providecommand{\natexlab}[1]{#1}
\providecommand{\url}[1]{\texttt{#1}}
\expandafter\ifx\csname urlstyle\endcsname\relax
  \providecommand{\doi}[1]{doi: #1}\else
  \providecommand{\doi}{doi: \begingroup \urlstyle{rm}\Url}\fi

\bibitem[alp(1984)]{alpha-conversion}
Conversion ({C}hapter 2).
\newblock In Barendregt, H.~P. (ed.), \emph{The Lambda Calculus}, volume 103 of
  \emph{Studies in Logic and the Foundations of Mathematics}, pp.\  22--49.
  1984.

\bibitem[Abbe et~al.(2023)Abbe, Bengio, Lotfi, and Rizk]{pmlr-v202-abbe23a}
Abbe, E., Bengio, S., Lotfi, A., and Rizk, K.
\newblock Generalization on the unseen, logic reasoning and degree curriculum.
\newblock In \emph{Proceedings of the 40th International Conference on Machine
  Learning}, volume 202 of \emph{Proceedings of Machine Learning Research},
  pp.\  31--60, 23--29 Jul 2023.

\bibitem[Azerbayev et~al.(2023)Azerbayev, Schoelkopf, Paster, Santos, McAleer,
  Jiang, Deng, Biderman, and Welleck]{LlemmaModel}
Azerbayev, Z., Schoelkopf, H., Paster, K., Santos, M.~D., McAleer, S.~M.,
  Jiang, A.~Q., Deng, J., Biderman, S., and Welleck, S.
\newblock Llemma: An open language model for mathematics.
\newblock \emph{ArXiv}, abs/2310.10631, 2023.

\bibitem[Baier \& Katoen(2008)Baier and Katoen]{Baier2008PrinciplesOM}
Baier, C. and Katoen, J.-P.
\newblock Principles of model checking.
\newblock 2008.

\bibitem[Clarke et~al.(2018)Clarke, Henzinger, Veith, and
  Bloem]{Clarke2018HandbookOM}
Clarke, E.~M., Henzinger, T.~A., Veith, H., and Bloem, R.
\newblock Handbook of model checking.
\newblock In \emph{Cambridge International Law Journal}, 2018.

\bibitem[Dargan et~al.(2020)Dargan, Kumar, Ayyagari, and Kumar]{dlSurvey2020}
Dargan, S., Kumar, M., Ayyagari, M.~R., and Kumar, G.
\newblock A {Survey} of {Deep} {Learning} and {Its} {Applications}: {A} {New}
  {Paradigm} to {Machine} {Learning}.
\newblock \emph{Archives of Computational Methods in Engineering}, 27\penalty0
  (4):\penalty0 1071--1092, September 2020.
\newblock ISSN 1886-1784.

\bibitem[Duret-Lutz et~al.(2022)Duret-Lutz, Renault, Colange, Renkin, Aisse,
  Schlehuber-Caissier, Medioni, Martin, Dubois, Gillard, and Lauko]{spot}
Duret-Lutz, A., Renault, E., Colange, M., Renkin, F., Aisse, A.~G.,
  Schlehuber-Caissier, P., Medioni, T., Martin, A., Dubois, J., Gillard, C.,
  and Lauko, H.
\newblock From {S}pot 2.0 to {S}pot 2.10: What's new?
\newblock In \emph{Proceedings of the 34th International Conference on Computer
  Aided Verification (CAV'22)}, volume 13372 of \emph{Lecture Notes in Computer
  Science}, pp.\  174--187, August 2022.

\bibitem[Frieder et~al.(2023)Frieder, Pinchetti, Griffiths, Salvatori,
  Lukasiewicz, Petersen, Chevalier, and Berner]{chatgpt-math}
Frieder, S., Pinchetti, L., Griffiths, R.-R., Salvatori, T., Lukasiewicz, T.,
  Petersen, P., Chevalier, A., and Berner, J.~J.
\newblock Mathematical capabilities of chatgpt.
\newblock \emph{ArXiv}, abs/2301.13867, 2023.

\bibitem[Grattafiori et~al.(2024)Grattafiori, Dubey, Jauhri, Pandey, Kadian,
  Al-Dahle, Letman, Mathur, Schelten, Vaughan, et~al.]{llama3herd}
Grattafiori, A., Dubey, A., Jauhri, A., Pandey, A., Kadian, A., Al-Dahle, A.,
  Letman, A., Mathur, A., Schelten, A., Vaughan, A., et~al.
\newblock The llama 3 herd of models.
\newblock \emph{arXiv preprint arXiv:2407.21783}, 2024.

\bibitem[Hahn et~al.(2021)Hahn, Schmitt, Kreber, Rabe, and Finkbeiner]{deepltl}
Hahn, C., Schmitt, F., Kreber, J.~U., Rabe, M.~N., and Finkbeiner, B.
\newblock Teaching temporal logics to neural networks.
\newblock In \emph{9th International Conference on Learning Representations,
  {ICLR} 2021, Virtual Event, Austria, May 3-7, 2021}, 2021.

\bibitem[Han et~al.(2021)Han, Rute, Wu, Ayers, and Polu]{Han2021ProofAC}
Han, J.~M., Rute, J.~M., Wu, Y., Ayers, E.~W., and Polu, S.
\newblock Proof artifact co-training for theorem proving with language models.
\newblock \emph{ArXiv}, abs/2102.06203, 2021.

\bibitem[Kamienny et~al.(2022)Kamienny, d'Ascoli, Lample, and
  Charton]{EndtoendSR}
Kamienny, P.-A., d'Ascoli, S., Lample, G., and Charton, F.
\newblock End-to-end symbolic regression with transformers.
\newblock \emph{ArXiv}, abs/2204.10532, 2022.

\bibitem[Lample \& Charton(2019)Lample and Charton]{Lample2019DeepLF}
Lample, G. and Charton, F.
\newblock Deep learning for symbolic mathematics.
\newblock \emph{ArXiv}, abs/1912.01412, 2019.

\bibitem[Li et~al.(2014)Li, Yao, Pu, Zhang, and He]{aalta}
Li, J., Yao, Y., Pu, G., Zhang, L., and He, J.
\newblock Aalta: an ltl satisfiability checker over infinite/finite traces.
\newblock In \emph{Proceedings of the 22nd ACM SIGSOFT International Symposium
  on Foundations of Software Engineering}, FSE 2014, pp.\  731–734, New York,
  NY, USA, 2014.

\bibitem[Liu et~al.(2023)Liu, Ash, Goel, Krishnamurthy, and
  Zhang]{liu2023transformers}
Liu, B., Ash, J.~T., Goel, S., Krishnamurthy, A., and Zhang, C.
\newblock Transformers learn shortcuts to automata.
\newblock 2023.

\bibitem[Morazzoni et~al.(2023)Morazzoni, Scotti, and Tedesco]{def2vec}
Morazzoni, I., Scotti, V., and Tedesco, R.
\newblock Def2vec: Extensible word embeddings from dictionary definitions.
\newblock In \emph{International Conference on Natural Language and Speech
  Processing}, 2023.

\bibitem[Pnueli(1977)]{Pnueli77}
Pnueli, A.
\newblock The temporal logic of programs.
\newblock In \emph{18th Annual Symposium on Foundations of Computer Science,
  Providence, Rhode Island, USA, 31 October - 1 November 1977}, pp.\  46--57,
  1977.

\bibitem[Press \& Wolf(2016)Press and Wolf]{weight-tying}
Press, O. and Wolf, L.
\newblock Using the output embedding to improve language models.
\newblock In \emph{Conference of the European Chapter of the Association for
  Computational Linguistics}, 2016.

\bibitem[Rabe et~al.(2020)Rabe, Lee, Bansal, and
  Szegedy]{Rabe2020MathematicalRV}
Rabe, M.~N., Lee, D., Bansal, K., and Szegedy, C.
\newblock Mathematical reasoning via self-supervised skip-tree training.
\newblock \emph{arXiv: Learning}, 2020.

\bibitem[Ranjan et~al.(2017)Ranjan, Castillo, and Chellappa]{l2face}
Ranjan, R., Castillo, C.~D., and Chellappa, R.
\newblock L2-constrained softmax loss for discriminative face verification,
  2017.

\bibitem[Sanh et~al.(2022)Sanh, Webson, Raffel, Bach, Sutawika, Alyafeai,
  Chaffin, Stiegler, Raja, Dey, Bari, Xu, Thakker, Sharma, Szczechla, Kim,
  Chhablani, Nayak, Datta, Chang, Jiang, Wang, Manica, Shen, Yong, Pandey,
  Bawden, Wang, Neeraj, Rozen, Sharma, drea Santilli, F{\'e}vry, Fries, Teehan,
  Scao, Biderman, Gao, Wolf, and Rush]{Sanh2021MultitaskPT}
Sanh, V., Webson, A., Raffel, C., Bach, S.~H., Sutawika, L., Alyafeai, Z.,
  Chaffin, A., Stiegler, A., Raja, A., Dey, M., Bari, M.~S., Xu, C., Thakker,
  U., Sharma, S.~S., Szczechla, E., Kim, T., Chhablani, G., Nayak, N.~V.,
  Datta, D., Chang, J., Jiang, M. T.-J., Wang, H., Manica, M., Shen, S., Yong,
  Z.-X., Pandey, H., Bawden, R., Wang, T., Neeraj, T., Rozen, J., Sharma, A.,
  drea Santilli, A.-., F{\'e}vry, T., Fries, J.~A., Teehan, R., Scao, T.~L.,
  Biderman, S., Gao, L., Wolf, T., and Rush, A.~M.
\newblock Multitask prompted training enables zero-shot task generalization.
\newblock In \emph{International Conference on Learning Representations}, 2022.
\newblock URL \url{https://api.semanticscholar.org/CorpusID:276421109}.

\bibitem[Shiv \& Quirk(2019)Shiv and Quirk]{treepos}
Shiv, V.~L. and Quirk, C.
\newblock Novel positional encodings to enable tree-based transformers.
\newblock In \emph{{NeurIPS} 2019}, December 2019.

\bibitem[Su et~al.(2024)Su, Ahmed, Lu, Pan, Bo, and Liu]{rope}
Su, J., Ahmed, M., Lu, Y., Pan, S., Bo, W., and Liu, Y.
\newblock Roformer: Enhanced transformer with rotary position embedding.
\newblock \emph{Neurocomput.}, 568\penalty0 (C), February 2024.
\newblock ISSN 0925-2312.
\newblock \doi{10.1016/j.neucom.2023.127063}.
\newblock URL \url{https://doi.org/10.1016/j.neucom.2023.127063}.

\bibitem[Tang et~al.(2023)Tang, Zheng, Li, Meng, Zhu, Liang, and
  Zhang]{Tang2023LargeLM}
Tang, X., Zheng, Z., Li, J., Meng, F., Zhu, S.-C., Liang, Y., and Zhang, M.
\newblock Large language models are in-context semantic reasoners rather than
  symbolic reasoners.
\newblock \emph{ArXiv}, abs/2305.14825, 2023.

\bibitem[Vastl et~al.(2022)Vastl, Kulhánek, Kubalík, Derner, and
  Babuška]{symformer}
Vastl, M., Kulhánek, J., Kubalík, J., Derner, E., and Babuška, R.
\newblock Symformer: End-to-end symbolic regression using transformer-based
  architecture, 2022.

\bibitem[Vaswani et~al.(2017)Vaswani, Shazeer, Parmar, Uszkoreit, Jones, Gomez,
  Kaiser, and Polosukhin]{vaswani}
Vaswani, A., Shazeer, N., Parmar, N., Uszkoreit, J., Jones, L., Gomez, A.~N.,
  Kaiser, L.~u., and Polosukhin, I.
\newblock Attention is all you need.
\newblock In \emph{Advances in Neural Information Processing Systems},
  volume~30, 2017.

\bibitem[Vazquez-Chanlatte \& Rabe()Vazquez-Chanlatte and Rabe]{pyaiger}
Vazquez-Chanlatte, M. and Rabe, M.
\newblock {py-aiger}.
\newblock URL \url{https://github.com/mvcisback/py-aiger}.

\bibitem[Wang et~al.(2017)Wang, Xiang, Cheng, and Yuille]{normface}
Wang, F., Xiang, X., Cheng, J., and Yuille, A.~L.
\newblock Normface: L2 hypersphere embedding for face verification.
\newblock In \emph{Proceedings of the 25th ACM international conference on
  Multimedia}, MM ’17, October 2017.

\bibitem[Wei et~al.(2016)Wei, Chen, Yang, Cao, and Zhang]{sign-lang}
Wei, S., Chen, X., Yang, X., Cao, S., and Zhang, X.
\newblock A component-based vocabulary-extensible sign language gesture
  recognition framework.
\newblock \emph{Sensors (Basel, Switzerland)}, 16, 2016.

\bibitem[Welleck et~al.(2022)Welleck, Liu, Lu, Hajishirzi, and
  Choi]{NaturalProver}
Welleck, S., Liu, J., Lu, X., Hajishirzi, H., and Choi, Y.
\newblock Naturalprover: Grounded mathematical proof generation with language
  models.
\newblock \emph{ArXiv}, abs/2205.12910, 2022.

\bibitem[Wu et~al.(2023)Wu, Qiu, Ross, Aky{\"u}rek, Chen, Wang, Kim, Andreas,
  and Kim]{reasoning-or-reciting}
Wu, Z., Qiu, L., Ross, A., Aky{\"u}rek, E., Chen, B., Wang, B., Kim, N.,
  Andreas, J., and Kim, Y.
\newblock Reasoning or reciting? exploring the capabilities and limitations of
  language models through counterfactual tasks.
\newblock In \emph{North American Chapter of the Association for Computational
  Linguistics}, 2023.

\bibitem[Yang et~al.(2023)Yang, Swope, Gu, Chalamala, Song, Yu, Godil, Prenger,
  and Anandkumar]{LeanDojo}
Yang, K., Swope, A.~M., Gu, A., Chalamala, R., Song, P., Yu, S., Godil, S.,
  Prenger, R.~J., and Anandkumar, A.
\newblock Leandojo: Theorem proving with retrieval-augmented language models.
\newblock \emph{ArXiv}, abs/2306.15626, 2023.

\bibitem[Zhang et~al.(2019)Zhang, Zhao, Qiao, Wang, and Li]{adacos}
Zhang, X., Zhao, R., Qiao, Y., Wang, X., and Li, H.
\newblock Adacos: Adaptively scaling cosine logits for effectively learning
  deep face representations.
\newblock \emph{2019 IEEE/CVF Conference on Computer Vision and Pattern
  Recognition (CVPR)}, pp.\  10815--10824, 2019.

\end{thebibliography}
